\documentclass{elsarticle}
\usepackage{amsfonts}
\usepackage{amsmath}
\usepackage{lineno,hyperref}

\usepackage{tablefootnote}
\usepackage{enumitem}
\usepackage{soul}
\setlist{label*=(\arabic*)}

\modulolinenumbers[5]

\journal{Journal}

\usepackage{numcompress}\biboptions{authoryear}

\begin{document}

\begin{frontmatter}

\title{Unsupervised Sentence Representations as Word Information Series: Revisiting TF--IDF}

\author[iifoot]{Ignacio Arroyo-Fern\'andez\corref{mycorrespondingauthor}\fnref{mymainaddress}}
\cortext[mycorrespondingauthor]{Corresponding author}
\ead{iaf@ciencias.unam.mx}
\address[mymainaddress]{Universidad Nacional Aut\'onoma de M\'exico (UNAM)}

\author[ccgfoot]{Carlos-Francisco M\'endez-Cruz\fnref{ccgaddress}}
\ead{cmendezc@ccg.unam.mx}
\address[ccgaddress]{Centro de ciencias gen\'omicas (CCG--UNAM)}
\fntext[ccgfoot]{Av. Universidad s/n Col. Chamilpa 62210, Cuernavaca, Morelos}

\author[iifoot]{Gerardo Sierra\fnref{iiaddress}}
\ead{gsierram@ii.unam.mx}
\address[iiaddress]{Instituto de ingenier\'ia (IIngen--UNAM)}
\fntext[iifoot]{AV. Universidad No. 3000, Ciudad universitaria, Coyoac\'an 04510, Ciudad de M\'exico}

\author[uapvfoot]{Juan-Manuel Torres-Moreno\fnref{uapvaddress}}
\ead{juan-manuel.torres@univ-avignon.fr}
\address[uapvaddress]{Laboratoire Informatique d'Avignon (LIA--UAPV)}
\fntext[uapvfoot]{Universit\'e d'Avignon et des Pays de Vaucluse. 339 chemin des Meinajaries 84911, Avignon cedex 9, France}

\author[cicfoot]{Grigori Sidorov\fnref{cicaddress}}
\ead{sidorov@cic.ipn.mx}
\address[cicaddress]{Centro de investigaci\'on en computaci\'on (CIC--IPN)}
\fntext[cicfoot]{Instituto Polit\'ecnico Nacional. Av. Juan de Dios B\'atiz, Esq. Miguel Oth\'on de Mendiz\'abal, Col. Nueva Industrial Vallejo, Gustavo A. Madero 07738, Ciudad de M\'exico}

\begin{abstract}
Sentence representation at the semantic level is a challenging task for Natural Language Processing and Artificial Intelligence. Despite the advances in word embeddings (i.e. word vector representations), capturing sentence meaning is an open question due to complexities of semantic interactions among words. In this paper, we present an embedding method, which is aimed at learning unsupervised sentence representations from unlabeled text. We propose an unsupervised method that models a sentence as a weighted series of word embeddings. The weights of the word embeddings are fitted by using Shannon's word entropies provided by the Term Frequency--Inverse Document Frequency (TF--IDF) transform. The hyperparameters of the model can be selected according to the properties of data (e.g. sentence length and textual gender). Hyperparameter selection involves word embedding methods and dimensionalities, as well as weighting schemata. 
Our method offers advantages over existing methods: identifiable modules, short-term training, online inference of (unseen) sentence representations, as well as independence from domain, external knowledge and language resources. 
Results showed that our model outperformed the state of the art in well-known Semantic Textual Similarity (STS) benchmarks. Moreover, our model reached state-of-the-art performance when compared to supervised and knowledge-based STS systems. 
\end{abstract}


\end{frontmatter}


\section{Introduction}
Nowadays, the growth of information in digital media encourages the analysis of large amounts of text data. This is attracting attention from Data Science and Artificial Intelligence researchers, as well as from the Internet industry. Internet users are responsible for a meaningful part of this growth. They enter information into the network which is also leveraged for sharing knowledge. An important part of this knowledge is found at repositories such as question \& answer forums, digital newspapers and digital encyclopedias.

Due to the innumerable duplication of the information at these repositories, several concerns arise as to the way users feed and consume knowledge. Some of these concerns include removing redundancies in question-answering forums or exploiting redundancies to assess the confidence of news in media or simply to compress text size. The accomplishment of this massive information processing is clearly infeasible for human reviewers. In this scenario, Statistical Natural Language Processing (NLP) methods are a substantial aid. 

An approach to address these issues is to perform massive comparisons by considering the content of sentences or short snippets of text. These comparisons can be done by means of Semantic Textual Similarity (STS) systems \citep{hatzivassiloglou1999detecting,agirre2012semeval}. An STS system computes a similarity score (a real value) between a pair of sentences. This score indicates how similar the sentences of the pair are. Most STS systems incorporate a number of supervisory signals such as Knowledge Bases, encyclopedias, language resources (e.g. thesaurus and linguistic taggers built on the basis of a Part-of-Speech [PoS] tagger) and even similarity labels \citep{mihalcea2006corpus}. Nevertheless, for specialized texts (or for low-resourced languages) those resources are not available or are scarce. Furthermore, the scope of such systems is limited exclusively to the task of measuring textual similarity. Such a limitation generally obviates the step of vector representation of sentences. This is not desirable when we want to study statistical behavior of sentence meaning.

Another approach is the one followed in this work. It consists of embedding sentences (or short text snippets) onto vector spaces such that approximations to their meanings can be represented geometrically, i.e. sentence embeddings or sentence representations \citep{hinton1986distributed,elman1991distributed}. The main advantage of such an approach is that there exists the possibility of studying the statistical behavior of sentence meaning. As an additional and important benefit, sentence embeddings make it possible to leverage a number of NLP tasks, such as sentence clustering, text summarization \citep{zhang2012mutual,arroyo2015learning,arroyo-fernandez-kdir16,yu2017learning}, sentence classification \citep{Kalchbrenner2014,chen2017improving,er2016attention}, paraphrase identification \citep{Yin2015DiscriminativePE}, semantic similarity/ relatedness and sentiment classification \citep{arroyofernandez-mezaruiz2017SemEval,chen2017improving,de2016representation,Kalchbrenner2014,onan2017hybrid,yazdani2013computing}.

The usefulness of vector representation methods mainly depends on the characteristics of the text to be embedded into vector spaces \citep{salton1988term}. On the one hand, most embedding methods provide well-suited representations of the content of texts for which individual size is on a relatively large scale \citep{salton1983extended,martin2007mathematical,mikolov2014prhase}. For instance, book-sized texts (e.g. documents with hundreds of thousands of words or larger) are well represented by the importance of the words they contain \citep{sparkJones1972}. Accordingly, the size of these text objects (as a Bag-of-Words, BoW) suggests that by representing them we can satisfy shallow information necessities limited to the gist (topics) of the documents \citep{manning2009,kintsch2011construction}, e.g. Information Retrieval and document classification. 

On the other hand, words are at the bottom end of the text size scale. At the word level, information necessities can be very general. That is, word representation methods could be components of practically any NLP system. These methods are mainly based on a general principle called \textit{distributional hypothesis}, which states that similar words are used in similar contexts \citep{Firth1957,harris1968}. In the NLP area, this linguistic principle is usually implemented as statistical estimates of word co-occurrence, i.e. word embedding methods. These statistical estimates provide word embeddings performing well enough in general purpose NLP applications \citep{baroni2010,mikolov2013efficient,pennington2014glove,baroni2014don,bojanowski2016enriching}. 

The problem of modeling sentences is still open. For the cases of documents or words, most applications expect representations encoding text content or word use. Nonetheless, in the case of sentences, application users can expect composite representations providing much more specific information, e.g. what is declared or denied about something \citep{pereira2000formal,meza2009jointly,collobert2011natural,kintsch2011construction}. Thus, state-of-the-art sentence representation methods can be highly dependent on the application and on its specificity. So it is difficult to keep their performance and behavior uniform/stable in several scenarios \citep{nghia2015jointly,pgj2017unsup}. Advancing the state of the art on sentence representation can be specially useful when only unlabelled text is available for learning sentence representations or even for applications to low-resourced languages.

In this work we address the problem of sentence representation by means of the following hypothesis. It is possible to represent sentences well by exploiting the link between the contexts learned by word embeddings and the entropy of such embedded words. In order to confirm our hypothesis, we present a modular model that consists of a weighted series of pretrained word embeddings. In this framework, the weights of the word embeddings are fitted by using unsupervised learning based on Shannon's entropy \citep{shannon1949communication}, i.e. TF--IDF (Term Frequency--Inverse Document Frequency). Thus, we take into account word importance, both at corpus level and at sentence level. 

We evaluated our model, called \textit{Word Information Series for Sentence Embedding} (\textit{WISSE}), by using well-known STS benchmarks provided by the SemEval competition (i.e. SICK [Sentences Involving Compositional Knowledge, 2014] and SemEval [2016]). Our results showed that WISSE outperformed (or was comparable to) strong state-of-the-art methods in such benchmarks. Additional advantages were observed, which were mainly due to the modularity and low computational cost of our model: short-term training, independence from domain, external knowledge and language resources, as well as online inference of (unseen) sentence representations.

The rest of this paper is organized as follows: Section \ref{sec:2relatedWork} presents the related work. Section \ref{sec:3distictionBrepresentationsAndSTS} presents the main differences between STS systems and sentence representation methods. Section \ref{sec:4motivationForWISSE} exposes the motivations for our method. Section \ref{sec:5theModulesOfWISSE} presents the modules composing our model. In Section \ref{sec:6wisse} we explain the constitution of our model and how the modules composing it interact. In Section \ref{sec:7designOfExperiments} we explain the design of our experiments and their objectives. Section \ref{sec:8bowwcbowResults} presents the obtained results and Section \ref{sec:9discussion} addresses the discussion about such results. Section \ref{sec:10possibleImprovements} provides insights on possible improvements, including advantages and disadvantages. Finally, in Section \ref{sec:conclusions} the conclusions derived from this work are presented.

\section{Related work}\label{sec:2relatedWork}
 The method proposed in this paper simultaneously falls into two categories of sentence representation methods, whose related work is reported in this section. The first category deals with models and methods that unsupervisedly embed short text snippets or sentences. Such an embedding is performed directly from unlabeled (plain) text data into vector spaces. The main methods are based on statistics and neural networks. The second category deals with models that use any form of weighted sum of word embeddings. This mechanism embeds text snippets or sentences into vector spaces by using pretrained word embeddings. In this framework the weights of the word embeddings can be fitted with both supervised and unsupervised learning.

\subsection{Unsupervised methods for sentence representation}

In this subsection we briefly describe a number of unsupervised methods whose aim is to build sentence representations for general purposes. These methods do not use external resources or supervisory signals, some of the main traits of the model presented in this paper. 

\subsubsection{Statistical methods} 
A popular statistical representation method was originally used in Information Retrieval applications, by which documents were represented. The representation consists of the TF--IDF transform of document vectors such that their components are, mainly, word frequencies. Currently there are multiple heuristics for computing the components, e.g. word presence/absence (binary), smoothed logarithm, etc. \citep{salton1983extended,salton1988term}. The transform gives a \textit{term-document} matrix as a result. Due to its monotonic nature, this model assigns relatively high weights to rare (infrequent) words and relatively low weights to very frequent words \citep{sparkJones1972}. According to information theory, this schema weighs the information conveyed by each word of the vocabulary \citep{aizawa2003information,robertson2004}. The weighting method consists of a logarithmic rescaling of the frequency of each word within a document. At the end, this logarithm linearizes the exponential distribution of word types through the corpus. In our experiments, we included BoW for sentence representation as a baseline method.

Latent Semantic Analysis (LSA) takes as its input a term-document matrix created by means of the BoW method \citep{landauer1998}. The sparse vectors of this matrix are transformed into document vectors, the of which are the projection weights to a user-defined number of eigenvectors of the term-document matrix. The transformation is computed by means of the Singular Value Decomposition (SVD) method. The number of eigenvectors is associated with the number of topics supposed to be present in the collection of documents \citep{martin2007mathematical}. 

\subsubsection{Neural sentence representations} 

The method called \textit{Doc2Vec} uses a neural network to build sentence/paragraph vectors that can be used for general purposes \citep{mikolov2014prhase}. This method uses word embeddings previously learned from fixed-length segments of text (sliding word windows). The words of a sentence are associated with the corresponding word embeddings \citep{mikolov2013distributed}. These embeddings then are used as evidence to predict a virtual word embedding which does not represent a word, but a sentence instead. 

As an extension of Doc2Vec, the neural model proposed by \cite{kiros2015skip} produces sentence embeddings from the hidden states of a Recurrent Neural Network (RNN). In this framework (the Skip-thought vectors), the two sentences surrounding a center sentence are a context window. The RNN maps these sentence contexts to its last hidden state, which is taken as a sentence embedding. 

The architecture called \textit{FastSent}, proposed by \cite{hill2016learning}, is based on the Glove model \citep{pennington2014glove} (see Section \ref{sec:wordEmbeddings}). Furthermore, the authors instantiate the architecture of Skip-Thought vectors \citep{kiros2015skip}. This combined network uses a precomputed matrix that merges co-occurrence information from both words and sentences. Additionally, \cite{hill2016learning} addresses the STS problem as one of machine translation. This helps to learn sentence representations by simulating negative examples to a Sequential (Denoising) Auto Encoder, S(D)AE, which uses such examples to learn a negative model. Thus, unseen sentences can be built in opposition to the jointly learned negative (adversarial) model.

The work of \cite{wieting-EtAl:2016:EMNLP2016} proposes a meaningful difference with respect to most word embedding methods. This model, called \textit{CHARAGRAM} learns embeddings of character $n$-grams \citep{bojanowski2016enriching}. Character embeddings are simply averaged in order to compose words. Actually the same operation is performed when sentence embeddings are needed, i.e. the obtained word embeddings are averaged to obtain sentence representations in the \textit{CHARAGRAM-PHRASE} model.

The neural sentence representation model called C-PHRASE relies on dependency/constituency parsing \citep{bentivogli2016sick}. The idea is very similar to the one proposed by \cite{levy2014dependency} for word embeddings (Section \ref{sec:dep2vec}). That is, the word co-occurrence is constrained by structure dependencies rather than by word-context windows scanning the input corpus. 

The model called \textit{Sent2Vec} is similar to that proposed by \cite{mikolov2014prhase}. The authors extended Doc2Vec for considering sentences, instead of fixed-length context windows. Additionally, this model can consider word $n$-grams or even the dynamic length of the context window as a modification of the subsampling approach proposed by \cite{mikolov2013efficient}. This model is also very similar to that proposed by \cite{bojanowski2016enriching} for word embeddings (Section \ref{sec:fastText}). That is, the architecture can learn a distribution of labels for a given training example.

Although sentence representation methods based purely on deep learning have shown competitive performance based on a number of benchmarks, their computational cost can be a significant bottleneck. Some of these methods need large amounts of data and even weeks of training on GPUs to perform reasonably well \citep{kiros2015skip}.

Most state-of-the-art methods are purely neural network based. Unlike such an approach, our method uses neural models only for training word embeddings (our word embedding module). 

\subsection{Weighted sum of word embedding methods}

The notion of weighting word embeddings composing sentences was introduced by \cite{ji2013discriminative}. A recent extension of such work is presented in \citep{Yin2015DiscriminativePE}. The authors appeal to supervised learning of weights for each word within a sentence. Their approach focuses on improving precision in paraphrase identification. This method rescales (enforced/penalized) the IDF of shared words of a pair of sentences. Herein, two complementary distributions of events are considered. First, the probability distribution that a given word taking place in sentences that are paraphrases. Second, the probability distribution that a given word taking place in sentences that are not paraphrases. The \textit{Kullback--Leibler} divergence of both distributions is computed to enforce IDF weights of shared word embeddings between paraphrases. The authors use negative factorization of the rescaled sentence matrices in order to build vector features for paraphrase classification.

In \citep{zheng2015learning,brokos2016using}, the authors improve Information Retrieval applications by using word embedding weighting. The embeddings of documents and queries are weighted via IDF. The goal is to compute the average similarity between query term embeddings and word embeddings of documents for retrieval. In \citep{kenter2015short} a similar approach is proposed. From partial measurements, a regression function is estimated from a training dataset in order to predict the rescaling of the word embeddings of a test set. 

In \citep{de2016representation}, the authors propose a supervised approach for learning the word embedding weights for text snippets (short texts). Their method represents snippets of paragraph-like lengths. A binary classification problem (relatedness/unrelatedness) is posed to learn the required weights. In this method, the IDF weights are used to rank the order of the weighted embedding summation. The authors show that unimportant words induce too much bias to semantic similarity measurement\footnote{We think that this noisy behavior is not necessarily due to the property of a word of being noisy, but more probably due to the over/underrepresentation issues in word embedding methods.}.  
Therefore their model drops unimportant words from the sentence representation. Thus their model learns uniquely the weights of surviving word embeddings. Next, the weighted word embeddings are averaged in order to obtain the final snippet representation. The authors pointed out that such an average works well for texts of about 30 words in length. Nonetheless, they also propose additional modifications to their model for variable text length. Unlike this approach, our study is for sentence representation.  

Another related method is the one recently proposed by \cite{ferrero-EtAl:2017:SemEval}. Their model uses PoS tag weights. These weights are merged to the associated IDF weights from the analyzed dataset of the STS sentence pairs. Both the PoS weights $\beta_{POS}$ and the IDF weights $\beta_{idf}$ of each word $w$ are combined using a product of powers. By means of such a combination, a unique word weight $\phi_w$ is obtained, i.e. $\phi_w=\beta_{POS}^{\alpha}\beta_{idf}^{1-\alpha}$. As both $\beta_{POS}$ and $\alpha$ are free parameters, the model learns them in a supervised fashion by using manually annotated semantic similarity scores. Thus, both word importance and word PoS tags contribute to weight word embeddings in order to build a sentence representation. 

In \citep{arora2017simple} the authors use word embedding weighting in both supervised and unsupervised approaches. The weights were learned as part of a multinomial distribution. This multinomial is parameterized according to the probability of a word appearing along with other words (i.e. the words of a sentence). This idea is actually an extrapolation of the principles behind distributed representations for word embeddings \citep{mikolov2013distributed}. However, this model additionally considers a balance (a linear convex compensation) between the probability of a word to occur within a \textit{discourse} and its probability to occur within a sentence.  

Most methods we presented in this subsection are supervised. An exception is the unsupervised method proposed by \cite{arora2017simple}, which is the most similar to ours. This method performs unsupervised learning of co-occurrence weights. Unlike to such an approach, ours performs unsupervised learning of weights based on information amount (which constitutes our information-theoretic module). 

\section{STS and the distinction between STS systems and sentence representation methods}\label{sec:3distictionBrepresentationsAndSTS}

Since we evaluated our sentence representation method by means of STS tasks, we briefly explore such tasks' context in this section. Furthermore, we show a comparison between sentence representation methods (like the proposed one in this paper) and STS systems. 

There are a number of proposed methods to assess semantic similarity between pairs of sentences. Most of these proposals emerge from benchmarks like the one used in the SemEval STS competition \citep{agirre2012semeval}. Given a pair of sentences, the aim of the STS task is to determine a similarity score (a real value). This score indicates just how similar the sentences of the pair are. The higher the score is, the higher the measured semantic similarity. 

STS benchmarks evaluate the correlation coefficient between the similarity assessed by some algorithm and the similarity assessed manually by humans. This coefficient (the Pearson's coefficient) is a real number $\rho(\cdot,\cdot)\in[-1,1]$. For instance, let $d=\{d(S_a^{(1)}, S_b^{(1)}),\dots,d(S_a^{(\ell)}, S_b^{(\ell)})\}$ be the similarities of a set of $\ell$ pairs of sentences. Also let $y=\{y_1,\dots,y_\ell\}$ be the gold standard of similarities manually annotated. In this example as $\rho(d,y)$ approaches $1.0$, it means that the STS algorithm producing $d$ performs well on the benchmark.

STS systems attaining relatively high performances have recently been developed. These systems correlate about $80\%$ with similarity scores annotated by humans \citep{cerEtAl2017}. Most systems integrate combinations of multiple algorithms providing partial scores from a number of aspects of the sentences \citep{sultan2014back}. For instance, a typical similarity score between sentences of a pair can be obtained as follows \citep{PILEHVAR201595,brychcin2016uwb}:
$$d(S_a, S_b)=\alpha_1d_1+\cdots+\alpha_nd_n.$$ 

In the $d(\cdot,\cdot)$ score, the $\alpha_i$s are weights controlling the influence of each aspect or attribute shared between sentences $S_a,S_b$. Such weights may be either user-defined or learned in a supervised fashion. Each $d_i$ is the overlapping score of the $i$th attribute (or category of attributes) aligned between sentences (e.g. words aligned by syntactic category, words aligned by their vicinity within a semantic graph, etc.). Notice that each $\alpha_id_i$ is actually a partial similarity score contributing independently to $d(\cdot,\cdot)$. 

There are also semisupervised systems whose main feature is the use of external resources such as knowledge bases, thesauruses and dictionaries \citep{PILEHVAR201595,kenter2015short,brychcin2016uwb}. Other methods also incorporate supervisory signals provided by PoS taggers, dependency/semantic parsers and Neural Networks \citep{rychalska2016samsung}. 

In our case, we evaluate the performance of our sentence representation method by using popular STS benchmarks, i.e. SemEval \citep{agirre2016semeval} and SICK \citep{bentivogli2016sick}. In this context, the evaluation of any sentence representation method like the ones shown in Section \ref{sec:2relatedWork} is very similar with respect to the evaluation of STS systems. Nonetheless, the interpretation of the results is quite different. 

When evaluating sentence representations, we only define a similarity function as an algorithm for comparing vector representations. For instance, the cosine similarity: $\cos(\theta)=\hat{y}(s_a, s_b)=(s_a\cdot s_b)/(\|s_a\|\cdot\|s_b\|)$. In this scenario, we should be focused on details involved in the construction of the representations. Thus, a set of similarities $\hat{y}=\{\hat{y}(s_a^{(1)}, s_b^{(1)}),\dots,\hat{y}(s_a^{(\ell)}, s_b^{(\ell)})\}$ is obtained from comparing a set of $\ell$ pairs of sentence representations. 

Let the gold standard of similarities manually annotated be $y=\{y_1,\dots,y_\ell\}$. As a good result on the STS benchmark we expect that the correlation coefficient to approach one, i.e. $\rho(\hat{y},y)\to 1.0$. It means that our sentence representation method generating the embeddings $s_{(\cdot)}\in\mathbb{R}^d$ performs well on the benchmark and with respect to the cosine similarity algorithm. Therefore, such representations encode a good approximation of the human sentence-meaning criterion given by $y$.

In general, STS systems do not needed to represent sentences at all. In fact, there is a remarkable distinction between the aim of an STS system and the aim of a sentence representation method evaluated in STS tasks. That is, most STS systems are designed to beat the STS ranking, while sentence embedding methods are designed to provide distributional semantic representations of sentences. See Table \ref{tab:systemsFeatures}. 

\begin{table}[!htbp]
\centering
\begin{scriptsize}
\caption{Comparison of characteristics between STS systems (\texttt{sts}) and sentence representation methods (\texttt{repr}). Column names: Access to sentence representations (\texttt{Access}=\{\texttt{Yes}/\texttt{No}\}), Partial similarity scores (\texttt{Partial}=\{\texttt{Yes}/\texttt{No}\}), use of external (knowledge) resources (\texttt{Resources}).}
\vspace{3mm}
\label{tab:systemsFeatures}
\begin{tabular}{p{3.2cm}|p{1.0cm}p{1.6cm}p{1.3cm}p{1.5cm}p{0.8cm}}
System/method & sts/ repr. &Parsing & Resources & Partial & Access \\
\hline
\cite{rychalska2016samsung}& sts &Dependency & WordNet & Supervised & No \\
\cite{han2013umbc} & sts &Chunking & WordNet & Yes & No \\
\cite{sultan2014back}& sts &Chunking, named entities, dependency & WordNet & Yes & No \\
UWBunsup \citep{brychcin2016uwb} & sts &Chunking & No & No & No \\
\hline
LSA                       & repr. & No & No & No & \textbf{Yes}\\
BoW                        & repr. & No & No & No & \textbf{Yes}\\
Sent2Vec & repr.  & No & no & No & \textbf{Yes}\\
Doc2Vec & repr.    & No & No & No & \textbf{Yes}\\
FastSent	& repr. & No & No & No & \textbf{Yes}\\
Skip-Thoughts & repr.  & No & No & No & \textbf{Yes}\\
Glove-WR & repr.  & No & No & No & \textbf{Yes}\\
C-PHRASE & repr.  & Constituent structure & No & No & \textbf{Yes}\\
\textbf{WISSE} & repr.     & No & No & No & \textbf{Yes}\\
\textbf{WISSE} + dependency based word embeddings& repr. & Dependency & No & No & \textbf{Yes}\\
\hline
\end{tabular}
\end{scriptsize}
\end{table}

In this paper, we focused our comparisons on unsupervised learning methods which require neither labeled data nor external resources. Moreover, notice that we focused on sentence representation methods rather than on STS systems. 

\section{Motivation for weighted series of word embeddings}\label{sec:4motivationForWISSE}

Currently there are multiple approaches for building sentence representations in vector spaces. Nonetheless, the complexity of the problem of keeping stable sentence representations through multiple scenarios remains a bottleneck in the NLP area. This is mainly because there exist linguistic and knowledge resources unique for a limited subset of information necessities and languages. So research on efficient unsupervised sentence representation methods can offer a promising approach to such a limitation. In this section we present the intuitions that motivated our contribution. These intuitions are given mainly in the sense of orthogonality of word embeddings as a desired trait of weighted series.

\subsection{Composition in distributional semantics}\label{sec:compositionality}
\cite{lapata2010} propose a number of candidate models for semantic composition that are empirically tested as heuristics yielding promising results. Among the candidate models, the asymmetric composition is particularly interesting for us. This model is a weighted sum of word embeddings. Its purpose is to approximate the meaning compositionality of short phrases (e.g. ``random variable'', ``meaning composition''). In this framework, the asymmetry is posed as the linguistic feature such that the \textit{head}[\textsc{h}] of a phrase must be more important than the \textit{dependent modifier}[\textsc{md}]. See examples \ref{ph:1} and \ref{ph:2}:
\begin{enumerate}[resume]
\item \textit{random}[\textsc{dm}] \textit{variable}[\textsc{h}]\label{ph:1}
\item \textit{meaning}[\textsc{dm}] \textit{composition}[\textsc{h}].\label{ph:2}
\end{enumerate}
The asymmetric composition model is given by:
\begin{equation}
\label{eq:composition}
p=\alpha x_{[\textsc{dm}]}+\beta x_{[\textsc{h}]},
\end{equation}
where it must be verified that $\alpha<\beta$. This inequality reflects the difference between the importance of the constituents (i.e. the word embeddings $ x_{[\textsc{dm}]},x_{[\textsc{h}]}\in\mathbb{R}^n$) of the resulting phrase $p\in\mathbb{R}^n$. According to \cite{Tian2017}, coefficients $\alpha,\beta$ are scalars drawn from a monotonic function. In this work, we consider that a reasonable choice for such a monotonic function is the Shannon's entropy \citep{shannon1949communication,charniak1996statistical,aizawa2003information}. The asymmetric model takes into account both word order and linguistic features determining the syntactic category of the resulting phrase embedding $p$. 

Since the composition approach (\ref{eq:composition}) is plain and natural, it has encouraged recent work. For instance, \cite{Tian2017} described theoretical conditions for the vector averaging operation as a model of composition in distributional semantics. Given word embeddings $ x_{[\textsc{dm}]}$ and $x_{[\textsc{h}]}$ that are geometrically uncorrelated, then making $\alpha=\beta=\frac{1}{2}$ actually causes $p$ to approach zero. In other words, the average operation causes embeddings of words co-occurring with low or moderated frequencies to cancel each other, so $p\to 0$. This effect suggests a linguistic intuition: uncorrelatedness (including orthogonality) between the meanings of words of a phrase occurs when the speaker constructs composite meanings. In contrast, as word meanings are correlated in context (e.g. words co-occurring frequently, like "cell phone"). Frequent co-occurrences lead to simple meanings that can easily be represented by means of the average between word embeddings. In this case, $p$ is actually encoded as an implicit embedding shared by $ x_{[\textsc{dm}]}$ and $x_{[\textsc{h}]}$.  From our experiments, we interpret that the aforementioned observations explain the low performance of the simple average of word embeddings for representing whole sentences (Section \ref{sec:8bowwcbowResults}). 

\subsection{The sparseness in neural language models}\label{sec:sparseness}  

Word embedding models have one characteristic in common: the sparseness of word co-occurrence statistics induces orthogonality \citep{elman1991distributed}. Now, we present the main cases during the training of word embeddings where orthogonality is relaxed and where it approximately holds. For instance in recent work on neural models \citep{bengio2003neural,mikolov2013efficient}, binary sparse vectors are used for representing input words as categorical variables (one-hot encoded vectors). These vectors build a canonical basis for $\mathbb{R}^{|V|}$. This basis encodes the vocabulary as an orthonormal set 
$$e=\{e_1,\dots,e_{|V|}\}=\left ( \begin{matrix}
1 & 0 & \cdots & 0\\ 
\vdots  & \ddots  & \cdots & \vdots \\ 
0 & \cdots & 0 & 1
\end{matrix} \right )$$ 
before training \citep{elman1991distributed}. After training, the projection layer of the neural model has learned a transformation relaxing the initial orthogonality of words co-occurring frequently. Conversely, for words that do not co-occur frequently, orthogonality is impregnated or imposed onto their associated embeddings. 

\subsection{The extreme values of training on co-occurrences}\label{sec:extremeValues}

Let us consider the case of Golve \citep{pennington2014glove}. Suppose that the word $w_i$ does not (or almost does not) co-occur along with the context $c_j=\{w_1,...,w_{i+r}\}$. Also, let $x_i$ and $\varphi_j$ be the corresponding embeddings. Such embeddings are learned by the Glove's loss:
\begin{equation}
\label{eq:gloveObjective}
\mathcal{J}(x_{i},\varphi_{j})=\sum_{i,j}^Vf(c_{ij})(\langle x_{i},\varphi_{j}\rangle-\log c_{ij})^2.
\end{equation}

If the word $w_i$ does not co-occur within the context $c_j$, then $c_{ij}\to 1$ and therefore $\log c_{ij}\approx 0$. As the objective (\ref{eq:gloveObjective}) requires that $(\langle x_{i},\varphi_{j}\rangle-0)^2\to 0$, then the dot product $\langle x_{i},\varphi_{j}\rangle\approx 0$. This implies that the word embeddings tend to be orthogonal. 

A different tendency can be observed when $w_i$ co-occurs frequently along with the context $c_j$. The training leads to $\log c_{ij} > b$, so that $\langle x_{i},\varphi_{j}\rangle\approx b$ for a sufficiently large $b>0$. Therefore the word embeddings are linearly dependent proportionally to $b$.  

Notice that this analysis considers the extreme values of co-occurrence. Depending on the stability of the embedding algorithms, orthogonality is uniquely a tendency that approximates an ideal distribution.

\subsection{Merging}
As an extension of the geometrical observations on compositionality and orthogonality, we propose weighted series of word embeddings for sentence representation. Weighted series can be seen as the decomposition of a sentence into basis vectors. Mathematically, a basis is an orthogonal set spanning a semantic space of sentences \citep{fourier1822theorie}. As being convenient for our proposed model, we decided to relax this concept. In our weighted series we include word embeddings just as produced by their algorithms (e.g. Word2Vec and Glove).  

\cite{pennington2014glove} offers a proof that the distribution of co-occurrences is bounded by a \textit{generalized harmonic number} (i.e. a set of logarithmic distributions). Such a proof and the analysis presented in sections \ref{sec:compositionality}, \ref{sec:sparseness} and \ref{sec:extremeValues} suggest that the angles between word embeddings are also logarithmically distributed as they are combined to transmit information. Thus, such a distribution relates Shannon's entropy, co-occurrence statistics and orthogonality. 
This reasoning in turn suggests that a weighted series model admits, on the one hand, the weighted contribution of word embeddings encoding composite meanings. On the other hand, this model also admits the weighted contribution of lexicalized phrases whose constituent words share implicit embeddings. Thus, theoretically a weighted series of word embeddings is built by two main subsets of word embeddings. The former subset contains a few uncorrelated (probably orthogonal) word embeddings carrying specific information. Also, the second (disjoint) subset contains correlated word embeddings carrying almost no information. Notice how this idea resembles the idea put forward in (\ref{eq:composition}).  

\section{The modules of our sentence representation method}\label{sec:5theModulesOfWISSE}

\subsection{The word embeddings module}\label{sec:wordEmbeddings}
In this section we will describe the word embedding methods that WISSE employs as one of its modules. One of the most popular methods is termed Word2Vec (W2V) \citep{mikolov2013efficient}, which is a neural language model inspired by ideas proposed previously \citep{hinton1986distributed,bengio2003neural}. Recently other methods have been proposed and they are still growing in popularity, e.g. Glove \citep{pennington2014glove} and FastText \citep{bojanowski2016enriching}. These methods share similar elements in general. Their aim is to train a learning model whose hidden parameters are random vectors. The model learns from multinomial distributions modeling the co-occurrence of each word type along with other words. These other words are encapsulated within sliding windows (i.e. target contexts for each word type \citep{Rong14}). Once the model is trained, its hidden parameters are used as word embeddings. 

\subsubsection{Word2Vec}

There are two possible neural architectures producing word embeddings with W2V: the Skip-gram and the CBoW. In this work, we used the Skip-gram neural architecture for training word embeddings. 

Let $V=\{w_1,\dots,w_t,\dots,w_{|V|}\}$ be the vocabulary of a context corpus $D$. The main idea behind Skip-gram is to (supervisedly) classify all word types $w_t$ as they are instantiated ($w_i=w_t$) though such a corpus $D=\{w_1,\dots,w_i,\dots,w_{|D|}\}$. Multiple target labels are simultaneously associated with each $w_t$. Each target set is the vocabulary of a sliding window having $r$ labels being the words around $w_i$. Thus, the target associated with $w_i$ takes the form $c_i=\{w_{i-r},\dots,w_{i+r}\}\setminus w_i$. The parameters of the distribution modeling the set $\{c_1,\dots,c_i,\dots,c_{|D|}\}$ must be learned by the Skip-gram classifier. 

Before training of the classifier, a preprocessing step transforms the input unlabeled text into a labeled one. This amounts to obtain a training set of the form $(w_1,c_1),\dots,(w_i,c_i),\dots,(w_{|D|},c_{|D|})$. As Skip-gram is actually a Neural Network, its output layer can be written in terms of its hidden layers as follows:
\begin{equation}\label{eq:softMax}
P(c_i|w_i=w_t)=P(w_{i-r},...w_{i+r}|w_t)= \frac{\exp(x_{t}, \varphi_{c_i})}{\sum_{w\in V}\exp(x_{t}, \varphi_w)}.
\end{equation}

The vectors $x_{t}\in \mathbb{R}^d$ are word embeddings of the word types $w_t$ and they are actually the columns of the first hidden layer of the Skip-gram. The dimension of the word embeddings is $d$, which equals the number of hyperplanes described by the first hidden layer. Entries of matrix $\varphi_{c_i}$ are the weights of the output layer, corresponding to the word types $w_{i-r},...,w_{i+r}\in c_i$. Entries of matrix $\varphi_{w}|_{w\in V}$ are all weights of the output layer. 

\subsubsection{Dependency-based word embeddings (Dep2Vec)}\label{sec:dep2vec}
One of the embedding methods we used as part of our experiments is an extension of W2V. 
The extension consists of a substantial modification of the notion of context window 
that uses dependency parsing as a surrogate for the standard co-occurrence window $c_i$ in (\ref{eq:softMax}). Given a word $w_i$, dependency parsing is used for establishing context restrictions. The pair $(w_{i\pm r}, w_i)$ is considered to co-occur within the same context whenever its words are related by a grammatical dependency. 
For example, according to the mentioned restrictions, the co-occurrence of the words \textit{John} and \textit{sneezes} has much higher probability than the co-occurrence of the words \textit{house} and \textit{sneezes}. \textit{House} and \textit{sneezes} can co-occur within a vicinity of words, but they do not hold a grammatical dependency. Therefore this pair is not considered as a co-occurring one. This is because of the object-subject distributional dependency restricting the actions that people and houses can do, i.e. a people sneeze, but a houses do not \citep{baroni2010,levy2014dependency,pgj2017unsup}.

\subsubsection{FastText}\label{sec:fastText}
 
The FastText model is another extension of W2V \citep{bojanowski2016enriching}. This model is a combination of the CBoW and Skip-gram architectures. FastText is designed to segment the input text into contiguous $n$-grams of characters $g_i$. These contiguous $n$-grams are then grouped into context windows $c_i$. As in the case of W2V, $c_i$ is associated with $w_i$ and $w_i\notin c_i$.  Thus, given a context window $c_i=\{g_1,...,g_{|c_i|}\}$ the model should predict a target window $c_i'=\{g_1,...,g_{|c_i|}, w_i\}$, i.e. $P(c_i'|c_i)=P(g_1,...,g_{|c_i|}, w_i|g_1,...,g_{|c_i|})$. Notice that both $c_i$ and $c_i'$ are a bit different than $c_i$ built in (\ref{eq:softMax}).  Furthermore, the target window contains the word $w_i$ which is composed by the $n$-grams $g_i$. This allows the model for different word embeddings for the word \textit{as} and for the bigram \textbf{as} composing the word \textit{where\textbf{as}}. The same mechanism
allows FastText to infer out-of-vocabulary word embeddings.  

\subsubsection{Global vectors for word representation (Glove)}
Global vectors for word representation (Glove) is a bit different from most of its counterparts. By bringing back pioneer methods like LSA, this model transforms a sparse matrix of word co-occurrences into co-occurrence probabilities  \citep{pennington2014glove}. The matrix of co-occurrences $C$ has entries $c_{ij}$ and each of them is the probability that the word $w_i$ co-occurs along with other words $w_j$ within a context window. The model estimates a regression function on the non-zero $\log$ probabilities. This considers as independent variables the set of word embeddings $x_{i}\in \mathbb{R}^d$. 

\begin{equation}\label{eq:glove}
\mathcal{J}(x_{i},\varphi_{j})=\sum_{i,j}^Vf(c_{ij})(\langle x_{i},\varphi_{j}\rangle-\log c_{ij})^2,
\end{equation}
where $x_{i}$ corresponds to the word $w_i$ and $\varphi_{j}\in\mathbb{R}^d$ is the embedding of the context window spanning over other words $w_j$. The weighting function $f(c_{ij})=\left(\frac{c_{ij}}{\max\{c_{ij}\}}\right)^\alpha$ allows $\mathcal{J}$ to be adapted to the importance of co-occurrence probabilities $c_{ij}$.\footnote{Similar considerations are also referred to as ``sub-sampling'' in \citet{mikolov2013efficient} and \citet{bojanowski2016enriching}.} Overall, the convex objective (\ref{eq:glove}) is conceived to optimize the correspondence between the inner product $\langle x_{i},\varphi_{j}\rangle$ and the log difference $\log c_{ij}= \log \frac{P(w_j|w_i)}{P(w_i)}$.

\subsection{The information-theoretic module}\label{sec:informationModule}

TF--IDF is a normalized log transformation providing three sources of information about the importance of words \citep{sparkJones1972,salton1983extended,aizawa2003information}. In our proposed model, these sources of information are merged so as to build a contribution function based on Shannon's entropy. 

Given a vocabulary $V$, the first source of information is the entropy of a sentence $s_j\in S$ due to the set of sentences $S$ (the corpus). The second source is the entropy of each word $w_i\in V$ due to the subset of sentences sharing it. The third source is the decrease of entropy of $S$ as $w_i$ is observed in a particular sentence $s_j\in S$.

Formally, according to \cite{aizawa2003information}, a sentence $s_j$ has probability $P(s_j)=1/N_S$ and all $s_j\in S$ are equally probable, so they are uniformly distributed. Now, it is defined the first information source as the entropy of each sentence $s_j$ due to $S$:
\begin{equation}\label{eq:idfa}
H(s_j)=-\sum_{s_j\in S}P(s_j)\log P(s_j) \approx-\log\frac{1}{N_S},
\end{equation}

\noindent where $N_S$ is the total number of sentences, i.e. $j=1,2,...,N_S$. 

In a similar way, we can define our second source of information. It is the entropy of a word $w_i$ due to the subset $\varsigma_i\subset S$ of sentences sharing such a word:

\begin{equation}\label{eq:idfb}
H(S|w_i)=-\sum_{s_j\in S}P(s_j|w_i)\log P(s_j|w_i)\approx-\log\frac{1}{N_{w_i}},
\end{equation}

\noindent where $N_{w_i}$ is the cardinality of $\varsigma_i$. One interpretation of the right side of (\ref{eq:idfb}) is that all elements of $\varsigma=\{\varsigma_i\subset S:i=1,2,\dots,|V|\}$ are equally likely and uniformly distributed.

Prior to computing our third information source we need to compute a conditional information source provided the first (\ref{eq:idfa}) and the second (\ref{eq:idfb}) information sources. It is the decrease in the entropy of $S$ and $\varsigma_i$ given that $w_i$ is observed \citep{Osteyee1974}. Such a decrease is also interpreted as the expected mutual information gained by $\varsigma_i$ and $S$ as each $w_i$ is observed:
\begin{equation}\label{eq:mutual}
 \begin{split}
 I(V,S) &= \sum_{w_i\in V} P(w_i)\left[ H(S)-H(S|w_i)\right]\\
 &\approx \sum_{w_i\in V} \frac{f_{w_i}}{F}\left(\log\frac{1}{N_{w_i}} -\log\frac{1}{N_S}\right).
 \end{split}
\end{equation}

The probability of any $w_i$ is computed as $P(w_i)=f_{w_i}/F$. $f_{w_i}$ is the frequency of $w_i$ in $S$ and $F$ is the sum of frequencies of all words in $S$. The conditional information described by (\ref{eq:mutual}) is a computation on the whole corpus. The marginal information amount associated with each word given a particular sentence can be obtained from each term of the summation (\ref{eq:mutual}). Such a marginal quantity is our third information source and it is defined explicitly in Section \ref{sec:6wisse}. 

\section{The proposed sentence representation model}\label{sec:6wisse}
In this paper, we propose to use the information provided by the entropies of words in a corpus (TF--IDF) as a weighting approach for series of word embeddings of the form 
\begin{equation}\label{eq:weigtedSeries}
s(x,\varphi)=\sum_{w_i\in s} \varphi_{i} x_{w_i}.
\end{equation}
In the general model (\ref{eq:weigtedSeries}), the unknown coefficients $\varphi_{i}$ weigh the contribution from each word embedding $x_{w_i}$ to the sentence representation $s(x,\varphi)$. A natural instinct is to optimize $\varphi_{1},\dots,\varphi_{|s|}$ without (or with few) assumptions about them. However, in this work we propose taking into account information features of each word obtained from a corpus as prior knowledge \citep{scholkopf1997prior}. 
This prior knowledge relies on the fact that topical meaningful words (e.g., noun phrases and named entities) have relatively low probability of being used \citep{robertson2004}, therefore they are informative. Conversely, non-topical words (e.g., prepositions and determinants) have a high probability of being used. Therefore, they are much less informative. Note that there is a correlation between syntactic features and how informative a word is \citep{mitra1997analysis,pereira2000formal,lapata2010,badarinza2017syntactic}.

Let us instantiate our information-theoretic module (Section \ref{sec:informationModule}):
\begin{equation}\label{eq:Mutual}
 I(V,S)\approx \sum_{w_i\in V} \frac{f_{w_i}}{F}\left(\log\frac{1}{N_{w_i}} -\log\frac{1}{N_S}\right).
\end{equation}
Now we use the prior knowledge it acquired once it has been trained on the corpus $S$. This means that, given the decomposition form of such a model, we can take the TF and the IDF vectors from it separately. Therefore the IDF vector $\varphi_S\in\mathbb{R}^{|V|}$ is given by:
\begin{equation}
\label{eq:idf}
\varphi_S= \left(\log\frac{1}{N_{w_1}},\dots,\log\frac{1}{N_{w_{|V|}}}\right) -\log\frac{1}{N_S} ,
\end{equation}
\noindent where $N_{w_i}$ is the number of sentences sharing each $w_i$ in (\ref{eq:Mutual}) and $N_S$ is the total number of sentences in $S$. Note that each component of $\varphi_S$ gives marginally the first and the second information sources given in Section \ref{sec:informationModule}.

Now, from each term $P(w_i)=\sum_{s_j\in S}P(w_i|s_j)\approx \sum_{s_j\in S}f_{ij}/F$ in (\ref{eq:Mutual}), we define $f_{ij}$ as the frequency of $w_i$ occurring within some sentence $s_j$. Therefore the probability of $w_i$ to occur in $s_j$ is given by $P(w_i|s_j)\approx f_{ij}/F$. Thus, we define our third information source. As a particular $w_i$ is observed in a particular $s_j$, it constitutes a conditional event decreasing the entropy of $S$ (which also increases its informativeness). Each term of (\ref{eq:Mutual}) therefore provides the informativeness of each particular event $(w_i,s_j)$, with which we define the TF vector:
\begin{equation}\label{eq:tf}
\varphi_{w_i}^{(s_j)} = \left(0,\dots,\frac{f_{ij}}{F} ,\dots,0\right)
\end{equation}

For consistency with $\varphi_S$, the TF vector $\varphi_{w_i}^{(s_j)}\in\mathbb{R}^{|V|}$ has a unique nonzero component $i$ in (\ref{eq:tf}). Now we merge our three information sources into the scalar weights $\varphi_{i}=\langle\varphi_{w_i}^{(s_j)},\varphi_S\rangle$ for each word embedding $x_{w_i}$ of the series (\ref{eq:weigtedSeries}), i.e. the associated TF--IDF weights. 

\begin{figure}
\centering
\includegraphics[scale=0.3]{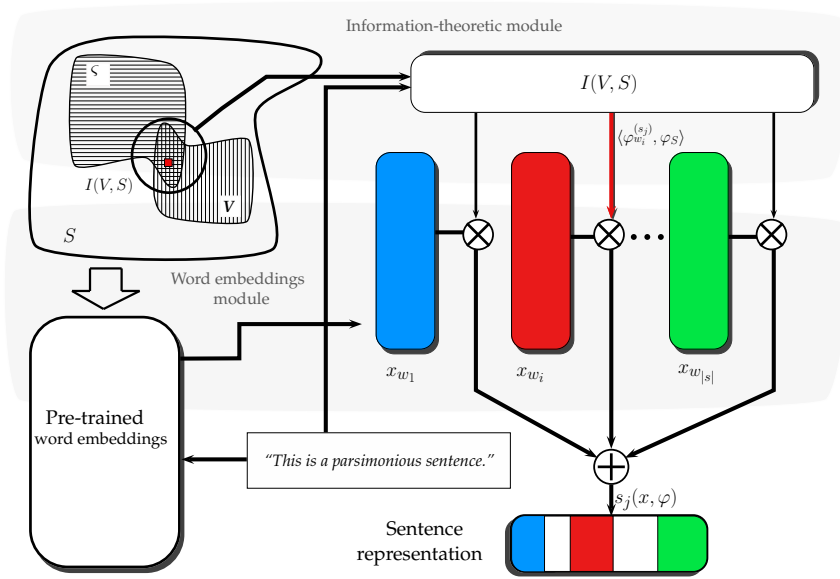}

\caption{Given a sentence (e.g. \textit{``This is a parsimonious sentence.''}), our model (\ref{eqSent}) asks to the word embedding module the word embeddings $\{x_{w_1},\dots,x_{w_{|s|}}\}$ associated with the words constituting such a sentence. In turn, the model asks to the information-theoretic module for the IDF vectors $\varphi_S$ and for the TF vectors $\varphi_{w_i}^{(s_j)}$ associated with each of these words. These vectors are multiplied, which gives us the TF--IDF weights $\langle \varphi_{w_i}^{(s_j)},\varphi_S\rangle$ for each word in the sentence. Once the needed TF--IDF values are obtained, the model assigns them as weights to their associated word embeddings. According to our model, the sentence representations $s_j(\varphi,x)$ are obtained simply by summing the weighted word embeddings.} 
\label{fig:sentenceRepresentationArch}
\end{figure}

Suppose that any of the models described in Section \ref{sec:wordEmbeddings} is already trained, e.g. via Glove (\ref{eq:glove}), so we can now instantiate our word embedding module. Let us pick up any word embedding $x_{w_i}$ from it. 

Let a sentence representation $s(\cdot,\cdot)$ be a vector derived from a textual construction whose elements (word embeddings) ``live'' in a semantic space $X\subset\mathbb{R}^d$. Now let $\varphi=(\varphi_{1},\dots,\varphi_{|s|}) =(\langle\varphi_{w_1}^{(s_j)},\varphi_S\rangle,\dots,\langle\varphi_{w_{|s|}}^{(s_j)},\varphi_S\rangle)\in\mathbb{R}^{|s_j|}$ be the coefficients of the weighted series (\ref{eq:weigtedSeries}), which have been learned by (\ref{eq:Mutual}) in an unsupervised fashion. Hence, we rewrite (\ref{eq:weigtedSeries}) as follows:
\begin{equation}\label{eqSent}
s_j(x,\varphi)=
\sum_{w_i\in s_j} \langle\varphi_{w_i}^{(s_j)},\varphi_S\rangle x_{w_i}.
\end{equation}
 
As a sentence representation $s_j(x,\varphi)$ is built, the learned vector $\varphi$ regulates the amount of information provided by each constituent word embedded in $x=\{x_{w_1},\dots,x_{w_{|s_j|}}\}$. See Figure \ref{fig:sentenceRepresentationArch}. We defined this regulatory process such that it is derived from three information sources. We have the constant vector $\varphi_S$ derived from the whole corpus $S$ and from a subset of sentences in it sharing each word $w_i$. Also we have the vector $\varphi_{w_i}^{(s_j)}$ 
which is derived from the subset of words $w_i\in V$ such that they take place in $s_j(\cdot,\cdot)$. This formulation proposes that word embeddings $x_{w_i}$ composing a sentence representation have different importances or contributions drawn by some abstract random process of communication, the language. 

\begin{figure}
\centering
\includegraphics[scale=1.0]{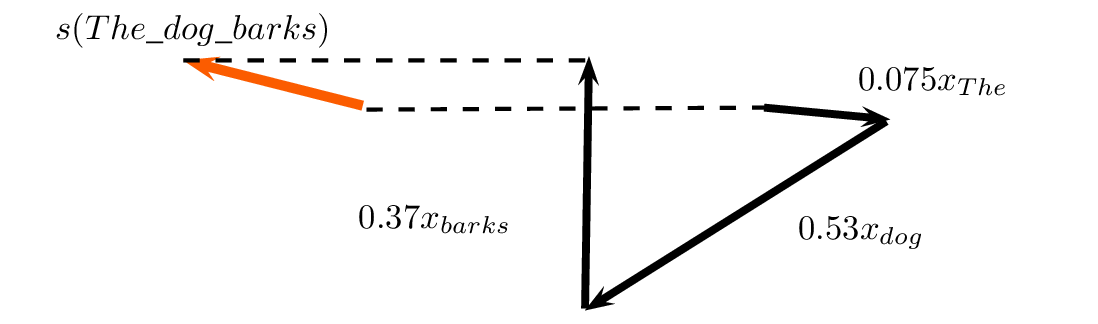}

\caption{A \textbf{hypothetical} sketch of the sentence vectors for the sentence \textit{``The dog barks''} and the actual weights computed from the corpus. The vector projected by the dotted lines (left side) is the hypothetical summation of constituents: the sentence representation.}
\label{fig:sentenceVector}
\end{figure}

As an example, we show a sketch of the sentence \textit{"The dog barks"}. Our model (\ref{eqSent}) allows to see how some $s(\cdot,\cdot)$ would look like geometrically (Figure \ref{fig:sentenceVector}). The sentence sketch could have weights like $\varphi_{The}=0.075, \varphi_{dog}=0.53, \varphi_{barks}=0.37$. Thus, the example $s[(x_{The},x_{dog},x_{barks}),\varphi]$ can be represented as
\begin{equation*}
\begin{split}
s(x,\varphi)=s[(x_{The},x_{dog},x_{barks}),\varphi]&=\sum_{w_i\in s} \langle\varphi_{w_i}^{(s)},\varphi_S\rangle x_{w_i}\\
                        &= 0.075x_{The} + 0.53 x_{dog} + 0.37x_{barks}
\end{split}
\end{equation*}

The modularity of our model allows for incorporating other information sources through its weights as required, e.g. syntax and structure \citep{tesitelova1992,de1999unsupervised,ferrero-EtAl:2017:SemEval}.

\paragraph{Time and memory cost} Suppose we have an NLP system that relies on sentence representations. Once we have pretrained word embeddings and entropy weights (e.g. TF--IDF), the sentence representations do not need to be explicitly inferred before they are needed. The representations can be computed \textit{online} as the application requires them. this is possible due to \textit{i)} the modularity of our model and \textit{ii)} the low-cost operations needed to embed a sentence representation. Given a sentence of $n$ words required by an NLP system, WISSE needs $d$ scalar multiplications to weight each word embedding. WISSE also needs the summation of the $n$ weighted word embeddings. Therefore the time needed for each sentence representation is $T_{mul}(dn)+T_{sum}(nd)=T_s(2n+2d)$. Given that the average sentence length is very short ($n\in\{8,12\}$ words), we see that $n\ll d$, so $T_s(2n+2d) \to \mathcal{O}(kd)$, which is a linear bound to the dimension of the word embeddings. Regarding the memory requirements, the elements of the model can be indexed (e.g. by a database). In this case, for each sentence we only need to load $n$ embeddings of $d$ dimensions and the corresponding $n$ weights, i.e. $n+nd=n(1+d)$. 

\section{Design of experiments}\label{sec:7designOfExperiments}
In this section we present the multiple aspects considered in designing our experiments. The datasets we used are described, followed by a description of the set of hyperparameters that we used for model selection.  

\subsection{Datasets}\label{sec:datasets}
 
We accomplished comparisons between our method and state-of-the-art methods on the basis of the SICK and the SemEval (2016) datasets \citep{bentivogli2016sick,agirre2016semeval}. Evaluations of some of these methods have been performed recently on the SemEval (2016) STS competitions \citep{king2016unbnlp,cerEtAl2017}. 

The SICK and the SemEval (2016) datasets collect sentence pairs from multiple sources. These datasets contain texts written by Internet users, so they are linguistically varied and raise a number of interesting challenges for research in unsupervised sentence representation methods. An important part of the datasets comes from repositories like forums and news appearing daily. The datasets also consider digital dictionaries and knowledge bases (e.g. Wikipedia, WordNet, FrameNet, OntoNet and others). In fact, the SICK dataset is a combination of carefully selected sentences from past years' SemEval STS datasets (2012-2014). 

In all cases, the pairs of sentences are associated with an averaged similarity gold standard score manually annotated. The score ranges from $0.0$, for unrelated pairs, to $5.0$, for equivalent or literally equal pairs. The details about the collection protocol of the dataset are described elsewhere \citep{agirre2016semeval,bentivogli2016sick}. We evaluated WISSE by taking advantage of this gold standard. A general summary of the datasets is shown below. 

\paragraph{The Answer-Answer dataset (SemEval-2016)} This dataset contains $1572$ pairs of answers which were extracted from the Stack Exchange Data Dump. The pairs of answers correspond to technical Stack Exchange forums, such as academia, cooking, coffee, DIY, and so on.

\paragraph{The Headlines dataset (SemEval-2016)} This dataset contains $1498$ pair of news headlines collected from the Europe Media Monitor (EMM).

\paragraph{The Postediting dataset (SemEval-2016)} This dataset contains $3287$ sentences from manually corrected machine translations of English-Spanish-French news. The translations were made using the Moses machine translation system. After that, the translations were manually post-edited so as to be corrected.

\paragraph{The Plagiarism dataset (SemEval-2016)} This dataset contains $1271$ short answers to computer science questions that exhibit varying degrees of plagiarism with respect to Wikipedia articles.

\paragraph{The Question-Question dataset (SemEval-2016)} This dataset contains $1555$ pairs of questions from the Stack Exchange Data Dump. The questions were extracted from technical Stack Exchange sites, such as academia, cooking, coffee, DIY, and so on. 

\paragraph{The OnWN dataset (SemEval-2013)} This dataset contains $561$ pairs of sense definitions of terms from WordNet and OntoNotes.                                                                                                                                                                                                                                                          
\paragraph{The FNWN dataset (SemEval-2013)} This dataset contains $189$ pairs of sense definitions of terms from WordNet and FrameNet.                                                                           

\paragraph{The SICK dataset (SemEval-2014)} This dataset contains $4906$ pairs of sentences selected from the STS Benchmark. This is a selection of the English datasets used in previous STS tasks (2012-2014). The selection is centered at the difficulty level in STS prediction. 

\paragraph{The Wikipedia dataset (general-purpose and unlabeled data)} This dataset contains $5.11$ million word types and it was downloaded from the Wikipedia dump (2012). 

\subsection{Experimental Setup}
The initial preparation for our experiments was to train the word embedding algorithms. For both W2V\footnote{\url{https://code.google.com/archive/p/word2vec}} and FastText\footnote{\url{https://github.com/facebookresearch/fastText}} word embedding models we used dimensions specified in Table \ref{tab:BOWwCBOWparameters}, as well as their usual training hyperparameters. The most important of them are the context window length (equal to $10$) and the minimum word frequency (equal to $2$). These word embedding models were trained on the Wikipedia dataset. For Glove we used $100$, $200$ and $300$ dimensions pretrained embeddings available at the authors' website\footnote{\url{https://nlp.stanford.edu/projects/glove}}. For Dep2Vec we used the $300-$dimensional pretrained embeddings available at the authors' website\footnote{\url{http://u.cs.biu.ac.il/~yogo/data/syntemb/deps.words.bz2}}.

With respect to information weights, we trained a TF--IDF model\footnote{\url{http://scikit-learn.org/stable/modules/feature_extraction.html}} on the Wikipedia for global training and on each STS dataset for local training. We used default settings, except that we explicitly specified in Table \ref{tab:BOWwCBOWparameters}.

So as to obtain a reliable idea of the possibilities of WISSE, we performed a number of experiments using datasets listed in Section \ref{sec:datasets}. To this end, we selected the hyperparameters of our model (WISSE). Their effects on the model are summarized in Table \ref{tab:BOWwCBOWparameters}. Combinations of such hyperparameters result in different versions of our model. Thus the idea is to select the best versions. 

Overall, both for SICK and SemEval (2016) datasets we first tuned the hyperparameters of our model (e.g. embedding dimension, TF--IDF weighting, and so on). State-of-the-art sentence representation methods have already been evaluated in previous work over these datasets (Section \ref{sec:datasets}). For each particular dataset we compared the Pearson coefficient obtained by state-of-the-art methods with that obtained by best versions of WISSE. This allowed us to compare the behavior of these methods, through varied scenarios.

\begin{table}[!htbp]
\centering
\begin{scriptsize}
\caption{Hyperparameters tuned to study WISSE's performance.}
\vspace{3mm}
\label{tab:BOWwCBOWparameters}
\begin{tabular}{l|p{5cm}|p{5cm}}
Name  & Description & Values \\
\hline
Dataset & The dataset for evaluation & \texttt{Plagiarism}, \texttt{Answer-Answer}, \texttt{Headlines}, \texttt{Postediting}, \texttt{Question-Question}, \texttt{FNWN}, \texttt{OnWN} and \texttt{SICK} (see Section \ref{sec:datasets})\\
\hline
Embedding & Word embedding method  & \texttt{W2V}, \texttt{Glove}, \texttt{FastText} and 
\texttt{Dep2Vec} \\
\hline
Size      & The embedding dimensions & \texttt{100d},\texttt{200d}, \texttt{300d}, \texttt{400d}, \texttt{500d}, \texttt{700d}, \texttt{1000d}.                                        \\
\hline
Comb      & Combination method to obtain each sentence representation & Summation (\texttt{sum}), average (\texttt{avg}).\\
\hline
Weights     & Weighting method giving word embedding coefficients. All weighting schemes were optionally computed with stop words removal. To denote it, the suffix \texttt{-st} was added. In the same way, the suffixes \texttt{-bin} and \texttt{-log} were added to denote whether the TF vector was computed either as binary or as the logarithm of word frequency. 
& 
\vspace*{-5mm}
\begin{list}{$\bullet$}{\leftmargin=1em \itemindent=0em}
\setlength\itemsep{0.1em}
\item TF--IDF global --with Wikipedia (\texttt{glob-tfidf}), 
\item IDF global (\texttt{glob-idf}),
\item TF--IDF local --with STS dataset (\texttt{loc-tfidf}), 
\item IDF local (\texttt{loc-idf}), 
\item All weights equal to \texttt{1.0} (unweighted).
\end{list} \\
\hline
Distance  & The distance to measure similarity between sentence vectors & \texttt{Cosine}, \texttt{Euclidean}, \texttt{Manhattan}.\\
\hline
\end{tabular}
\end{scriptsize}
\end{table}


In addition to the hyperparameters tuned for our model, we decided to test three different similarity functions for computing semantic similarity scores from WISSE sentence representations: the cosine similarity, the Euclidean distance and the Manhattan distance. For these three functions, the result is directly used as a predicted similarity score\footnote{As the Pearson's coefficient is scale-invariant it is not needed to transform the cosine similarities (which are $[-1.0,1.0]$) or Euclidean and Manhattan distances (which are $[0.0,\infty)$) into the range of similarities of the gold standard (which are $[0.0,5.0]$).} (Section \ref{sec:3distictionBrepresentationsAndSTS}). 

\section{Experimental Results}\label{sec:8bowwcbowResults}
Our results are divided mainly into two parts. We presented first the results of the experiments for the SICK dataset. At this stage, we tuned hyperparameters of the WISSE model. After that, we presented the comparison between our best result and the best results obtained by the state-of-the-art unsupervised sentence representation models. 

As in the case of SICK, we also tuned WISSE hyperparameters for the SemEval (2016) dataset. Accordingly, the best results obtained with WISSE were compared with the results obtained with state-of-the-art methods. Since the SemEval dataset comprises other datasets (Section \ref{sec:datasets}), we presented our best results obtained for each of the SemEval STS datasets (2016). 

Finally, we present a statistical comparison between our best result and the best results obtained by the state-of-the-art methods on the SemEval STS datasets. This last comparison includes not only unsupervised sentence representation models, but also supervised representation models as well as supervised and unsupervised STS systems.

\subsection{Hyperparameter tuning on the SICK dataset} 
The combination of all parameters of the model gave us a ranking of up to 200 experiments for the SICK dataset. The best 10 hyperparameter combinations for the three similarity functions (cosine, Euclidean and Manhattan) are shown in Table \ref{tab:sickWISSE}.  

\begin{table}[!htbp]
\centering
\begin{small}
\caption{WISSE hyperparameter combination results on the SICK dataset (train+test)}
\vspace{3mm}
\label{tab:sickWISSE}
\begin{tabular}{llllccc}
Weights        &Comb.    &Size  &Embedding  & Cosine $\rho$  & Euclid. $\rho$ & Manhatt. $\rho$  \\
\hline
 glob-tfidf-bin-st & sum     & 300d & FastText  & \textbf{0.72405} & 0.64465   & 0.64387  \\
glob-tfidf          & sum     & 200d & FastText  & 0.72023 & 0.64657   & 0.6469   \\
 glob-tfidf-bin      & sum     & 300d & W2V       & 0.71995 & 0.66747   & 0.66751  \\
 loc-tfidf-log      & sum     & 350d & FastText  & 0.71905 & 0.65583   & 0.65615  \\
loc-tfidf-bin      & sum     & 400d & FastText  & 0.71852 & 0.65236   & 0.65209  \\
loc-tfidf-st     & avg     & 300d & Glove     & 0.70397 & 0.61817   & 0.61885  \\
glob-tfidf-st     & sum     & 300d & Dep2Vec   & 0.67925 & 0.60972   & 0.61018  \\
loc-tfidf-log      & avg     & 300d & W2V       & 0.67428 & 0.6199    & 0.62075  \\
loc-tfidf-bin-st & avg     & 300d & W2V       & 0.66410 & 0.58308   & 0.58289  \\
loc-tfidf-log      & avg     & 300d & Dep2Vec   & 0.64762 & 0.55620   & 0.55585  \\
\hline
\multicolumn{7}{l}{\footnotesize Bold value indicates the best result.}\\
\end{tabular}
\end{small}
\end{table}
Our model reached its maximum correlation of $\rho=0.72405$ for the \texttt{cosine} similarity function. The IDF vector was globally learned (\texttt{glob}) from the Wikipedia corpus, which was also used for training the word embeddings. The TF vector was derived binary (\texttt{bin}) from the words in each sentence pair, i.e. $\varphi_{w_i}^{s_j}=\{ f_{ij}>0$ \texttt{?} $1.0/F : 0.0\}$. Moreover, in this particular experiment, we stripped out stop words (\texttt{-st}) and, therefore, they were omitted from the obtained sentence representations. \texttt{FastText} embeddings of $300$ dimensions (\texttt{300d} size) resulted in the best ones over \texttt{W2V}, \texttt{Glove} and \texttt{Dep2Vec}. The weighted word embeddings were combined by adding them (\texttt{sum}). We also performed experiments using a simple word embedding average (\texttt{avg}). It did not give us higher results than the state-of-the-art mean. Notice that for this dataset, no other similarity function was better than the \texttt{cosine}. Therefore, this dataset seems to be best characterized by the angles between word embeddings in sentence representations.

\subsection{Results for the SICK dataset}
The main result presented in this paper is the performance of our model with respect to the state of the art on unsupervised sentence representation methods.

\begin{table}[!htbp]
\centering
\begin{small}
\caption{Performance of sentence representation models on the SICK dataset}
\vspace{3mm}
\label{tab:overallSOA}
\begin{tabular}{lc}
Sentence representation method (reference)           & Pearson\\
\hline
Glove+WR \citep{arora2017simple}                     & 0.722    \\
Sent2vec \citep{pgj2017unsup}                          & 0.720    \\
FastSent \citep{hill2016learning}                      & 0.720    \\
C-PHRASE \citep{nghia2015jointly}                      & 0.720    \\
CHARAGRAM-PHRASE \citep{wieting-EtAl:2016:EMNLP2016}  & 0.700    \\
Skip-thoughts \citep{kiros2015skip}                   & 0.600    \\
BoW TF--IDF \citep{salton1983extended}                  & 0.580    \\
SDAE \citep{hill2016learning}                          & 0.460    \\
Doc2Vec \citep{mikolov2014prhase}                     & 0.460    \\
SAE \cite{hill2016learning}                           & 0.310    \\
\textbf{WISSE} (\texttt{glob-tfidf-bin-st,sum,300d,FastText}) & \textbf{0.724} \\
\hline
\multicolumn{2}{l}{\footnotesize Bold value indicates the best result.}\\
\end{tabular}

\end{small}
\end{table}

 One of the most popular tasks for evaluating unsupervised sentence representations is using them to predict similarity scores of the SICK dataset \citep{bentivogli2016sick}. Therefore, we compared the correlation given by our best model and the correlation given by state-of-the-art models (Table \ref{tab:overallSOA}). 

To our knowledge, both \cite{arora2017simple} and \cite{pgj2017unsup} are the best unsupervised sentence representation methods at this moment. Additionally we collected 
results obtained by other unsupervised methods and the Bag-of-Words baseline (which uses binary TF). Based on the Pearson's correlation coefficient, WISSE performed better than all them ($\rho=0.724$). 

It is important to notice that the difference is not large between WISSE and the following two methods in the ranking (Table \ref{tab:overallSOA}). There are now 5 sentence embedding methods within such a barrier (Glove+WR, Sent2vec, FastSent, C-PHRASE and WISSE). 

\subsection{Hyperparameter tuning on the SemEval STS datasets}\label{sec:resultsTuningSemeval}
We obtained correlation coefficients for WISSE on the SemEval STS datasets (2016) (see Table \ref{tab:wisseSTS}). These datasets together present much more variety with respect to the SICK dataset. Therefore, the results also varied independently of the evaluated model. 

Our best result was obtained for the Postediting dataset ($\rho=0.82161$). The \texttt{tfidf} weights were computed locally from the Postediting dataset by using word frequencies as TFs. Stop words were included in the model. The best embeddings for this dataset were Dep2Vec of 300 dimensions. The Euclidean distance better captures the similarity between sentence representations. Notice that there is a relatively large difference between the metrics (Euclidean and Manhattan) and the cosine similarity. Postediting was the unique dataset for which Dep2Vec embeddings led to the best result. 

Our second best result was obtained for the Plagiarism dataset ($\rho=0.80607$). For this dataset the best hyperparameters varied considerably with respect to the best for Postediting. The main differences were observed in word embedding weighting and word embeddings. The \texttt{tfidf} weights were computed globally from Wikipedia and by using binary TFs. The best word embeddings were W2V of 300 dimensions.

Our lowest result in SemEval (2016) was for the Answer-Answer dataset ($\rho=0.6556$). This dataset has been reported to be challenging for most STS competitors \citep{agirre2016semeval}. Our model needed W2V embeddings of 1000 dimensions. The weights of the model were computed locally and by using the logarithm of the word frequencies in the sentences, i.e. $\varphi_{w_i}^{(s_j)}=\log (f_{ij} +1 )/F$.

\begin{table}[!htbp]
\centering
\begin{small}
\caption{WISSE hyperparameter combination results on the SemEval STS dataset (2016)}
\vspace{3mm}
\label{tab:wisseSTS}
\begin{tabular}{l|llllll}
Dataset & Embedding & Size & Weights & Cosine   & Euclidean  &Manhatt.\\
\hline
Postediting & Dep2Vec&300d& loc-tfidf & 0.652880 &\textbf{0.821610}& 0.819890\\
Plagiarism & W2V & 300d & glob-tfidf-bin  & 0.775750& \textbf{0.806070}& 0.805280  \\
Ques.-Ques. & FastText  & 300d& glob-tfidf  & \textbf{0.704010} & 0.683410  & 0.681400  \\
Headlines & FastText  & 200d & glob-tfidf  & 0.676300 &\textbf{0.701020} & 0.700720 \\
Ans.-Ans. & W2V  & 1000d & loc-tfidf-log  & 0.507660 &\textbf{0.655600} & 0.652110\\
\hline
OnWN & W2V  & 1000d & loc-tfidf-log-st & \textbf{0.833070} &0.739430 & 0.738610\\
FNWN & FastText  & 200d   & glob-tfidf  & \textbf{0.458560} & 0.350300 & 0.363520  \\
\hline
\multicolumn{7}{l}{\footnotesize For all cases the best combination operation was the summation (\texttt{sum}).}\\
\multicolumn{7}{l}{\footnotesize Bold value indicates the best result.}\\
\end{tabular}
\end{small}
\end{table}

In most cases, the two metrics (Euclidean and Manhattan) outperformed the cosine similarity and the difference is relatively large. We observed this for the majority of the datasets (Table \ref{tab:wisseSTS}): Postediting, 
Plagiarism, Headlines and Answer-Answer. For all datasets, the weighted word embeddings were summed in order to obtain the corresponding sentence representation. In fact, all our best results were obtained by summing the weighted embeddings (there were no cases where the word embedding average (\texttt{avg}) performed better).

The dimension of the embeddings did not show specific patterns, except for the OnWN and Answer-Answer datasets. Both of them needed logarithmic TF vectors as well as $1000$ dimensional word embeddings. In contrast to the other datasets, these datasets allowed incremental dimensionality augmentation ($200-1000$). Nonetheless, once such an augmentation surpassed 500 dimensions the performance almost did not improve. For all other datasets, augmenting dimensions degraded the performance. WISSE performed better for all the datasets when the weights were the TF--IDF vector products (using either binaries, frequencies or their logarithm) instead of only IDF vector components. This observation could be considered independent of the word embedding method and dimension. 

As an additional test we evaluated our model on the OnWN, FNWN (SemEval 2013) datasets.
These datasets have been reported to be challenging \citep{agirre2013sem}. 

On the one hand the OnWN dataset was shown to be well represented by the BoW baseline, $\rho = 0.8431$; however for SemEval competing systems it is hard to surpass such a performance. Our best result was comparable ($\rho=0.82161$) by using W2V embeddings of $1000$ dimensions. The IDF vectors were locally learned and the TF vectors were computed from the logarithm of the word frequency in sentences. The stop words were stripped out.  

On the other hand, for the FNWN dataset the competing systems were struggling to reach the maximum $\rho=0.5818$. Our system reached $\rho=0.45856$ by using FastText word embeddings of $200$ dimensions. The embeddings were weighted via globally learned IDFs and frequency TFs. For both datasets, the cosine similarity worked considerably better than the metrics. 

\subsection{Results for the SemEval STS task} 

We compared WISSE with the sentence embedding methods from the state of the art on the SemEval STS dataset (2016). For WISSE, we have used the best hyperparameters fitted for each dataset (Section \ref{sec:resultsTuningSemeval}).

\begin{table}[!htbp]
\centering

\caption{WISSE and state-of-the-art sentence embedding models on SemEval STS datasets}
\vspace{3mm}
\label{tab:soaPosition}
\begin{scriptsize}
\begin{tabular}{l|ccccc}
 Model & Ans.-Ans. & HDL & Plagiarism & Postediting & Ques.-Ques. \\
\hline
Sent2Vec\tablefootnote{This method was tested by the MayoNLP team \citep{afzal2016mayonlp}
.} & 0.57739   & \textbf{\textit{0.75061}} & 0.80068    & \textbf{\textit{0.82857}}     & \textbf{\textit{0.73035}}     \\
\textbf{WISSE}    & \textbf{0.65560} & 0.70102 & \textbf{0.80607} & 0.82161 & 0.70410\\
D2V (400d) & 0.41123 & 0.69169 & 0.60488 & 0.75547 & -0.02245 \\
Skip-toughs & 0.23199 & 0.49643 & 0.48636 & 0.17749 &  0.33446\\
\hline
W2V (300d-average) & 0.50311 & 0.66362 & 0.72347 & 0.73935 & 0.16586 \\
Binary TF--IDF (BoW) & 0.41133 & 0.54073 & 0.69601 & 0.82615 & 0.03844 \\
\hline
\multicolumn{6}{l}{\scriptsize Bold values indicates the best result for WISSE.}\\
\multicolumn{6}{l}{\scriptsize Values in bold italics indicate best state-of-the-art results.}
\end{tabular}
\end{scriptsize}
\end{table}

We observed that for two datasets (Answer-Answer and Plagiarism) WISSE performed better than the state of the art methods (Table \ref{tab:soaPosition}). For the other three datasets (Headlines (HDL), Postediting and Question-Question) the Sent2vec method was better. We observed that, in general, it is difficult to beat the BoW baseline in the Postediting dataset. Sent2Vec surpassed it by a small difference. WISSE was lower, also, by a small difference. 

Although currently D2V (Paragraph vector) and Skip-Thoughts are very popular methods, their performances in STS were even lower than that of the BoW baseline (excepting the Postediting dataset). The performances of such methods were also considerably lower than those of Sent2Vec and WISSE. For instance, while Sent2vec and WISSE reached roughly $\rho>0.70$ on the Question-Question dataset, D2V fell to $\rho=0.02245$. A similar situation occurred for the Skip-Thoughts method, which reached $\rho=0.17749$ on the Postediting dataset while the best methods were $\rho>0.82$. Finally, the simple average of W2V word embeddings in building sentence representations showed to be a method with very similar performance with respect to results obtained with D2V.

\subsection{The SemEval STS task}\label{sec:semevalPosition}
We performed a comparison between WISSE and the top $10$ STS systems in the SemEval STS (2016) competition (Table \ref{tab:stsPosition}). Although we did not participate in such a competition, it is relevant to observe how WISSE performed with respect to state-of-the-art STS systems. Notice that these systems can be designed in varied forms and for different purposes than WISSE (Section \ref{sec:3distictionBrepresentationsAndSTS}). That is, while WISSE aims to unsupervisedly represent sentences in vector spaces, STS systems aim to measure semantic textual similarity (un/supervisedly and by using varied external resources). 
\begin{table}[!htbp]
\centering

\caption{Ranking for the SemEval STS 2016 task (this is a modified version derived from \cite{agirre2016semeval})}
\vspace{3mm}
\label{tab:stsPosition}
\begin{tiny}
\begin{tabular}{c|l|c|ccccc}
R & System & ALL & Ans.-Ans. & HDL & Plagiarism & Postediting & Ques.-Ques. \\
\hline
1 &Samsung Pol.  & 0.77807    & 0.69235   & 0.82749 & 0.84138    & 0.83516     & 0.68705     \\
2 &UWB & 0.75731    & 0.62148   & 0.81886 & 0.82355    & 0.82085     & 0.70199     \\
3 &MayoNLPTeam & 0.75607    & 0.61426   & 0.77263 & 0.805      & 0.8484      & 0.74705     \\
4 &Samsung Pol.  & 0.75468    & 0.69235   & 0.82749 & 0.81288    & 0.83516     & 0.58567     \\
5 &NaCTeM & 0.74865    & 0.60237   & 0.8046  & 0.81478    & 0.82858     & 0.69367     \\
6 &ECNU & 0.75079    & 0.56979   & 0.81214 & 0.82503    & 0.82342     & 0.73116     \\
7 &UMD-TTIC-UW & 0.74201    & 0.66074   & 0.79457 & 0.81541    & 0.80939     & 0.61872     \\
9 &SimiHawk & 0.73774    & 0.59237   & 0.81419 & 0.80566    & 0.82179     & 0.65048     \\
\hline
8 &Sent2Vec & 0.73836    & 0.57739   & 0.75061 & 0.80068    & 0.82857     & 0.73035     \\
10&\textbf{WISSE}        & 0.73768 & 0.655600 & 0.70102 & 0.80607 & 0.82065 & 0.70410\\
23&UWB & 0.72622    & 0.64442   & 0.79352 & 0.82742    & 0.81209     & 0.53383     \\
--&D2V (400d) & 0.50206 & 0.41123 & 0.69169 & 0.60488 & 0.75547 & -0.02245 \\
--&Skip-toughs & 0.27148 & 0.23199 & 0.49643 & 0.48636 & 0.17749 &  0.33446\\
\hline
--&W2V (300d-avg) & 0.56007 & 0.50311 & 0.66362 & 0.72347 & 0.73935 & 0.16586 \\
110&STS (BoW) & 0.51334 & 0.41133 & 0.54073 & 0.69601 & 0.82615 & 0.03844 \\
\hline 

\end{tabular}
\end{tiny}
\end{table}

We used our best hyperparameters for each dataset (Section \ref{sec:resultsTuningSemeval}). The general score for the competition is a weighted average of the correlations from all datasets (column \texttt{ALL}). It is encouraging that WISSE is ranked at $10$\textsuperscript{th} place among $113$ systems. This comparison was performed putting the characteristics of all those systems aside. This is an important consideration because most of them are supervised and include several lexical resources like WordNet, FrameNet and dictionaries. Thus, the unsupervised sentence representation models are separated at the middle of Table \ref{tab:stsPosition} in order to keep such a distinction.

\begin{figure}[!htbp]
\centering
\includegraphics[scale=0.5]{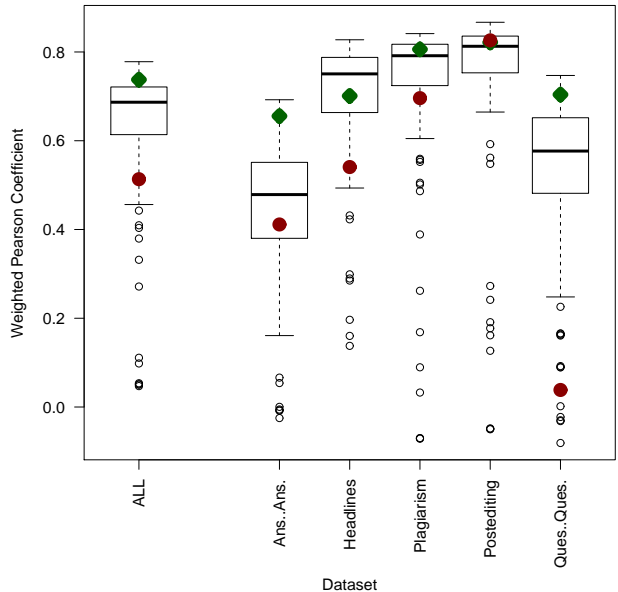}
\caption{Box plot of the statistical position of the WISSE model within the SemEval STS 2016 state of the art. The vertical axis is the correlation coefficient given by participant systems for each dataset of the STS competition. The horizontal axis shows the overall performances (\texttt{ALL}) and the performance for each dataset.
(WISSE = green diamonds; BoW baseline = red circles).}
\label{fig:positionSemeval}
\end{figure}

To know whether our model is actually competitive, we analyzed the correlation coefficients given by participant systems (Figure \ref{fig:positionSemeval}) for the overall performances (\texttt{ALL}) and the performance for each dataset (Table \ref{tab:stsPosition}). The first observation is that the WISSE model (green diamond) outperformed the mean for almost all datasets. For the Headlines dataset, WISSE did not surpass the mean, but it stayed within the inter quartile range (IQR). Neither the Plagiarism nor the Postediting datasets, do not present high difficulties for the state-of-the-art systems. This can be a result of the low variability and the high overall mean correlation reached by most systems for this dataset. In the same vein, the BoW baseline (red circle) overpasses the mean correlation for the Postediting dataset. For these datasets, our system remains within the IQR, but not too far away from the maximum correlation.

Statistically, our best results were reached on both the Answer-Answer and the Question-Question datasets. Our model was allocated beyond the first quartile, near to the maximum overall performances. That is, WISSE succeed in dealing with the uncertainty of the mean correlations reached by the 95\% of the systems (IQR) on such difficult datasets. For the Question-Question dataset the baseline is at the lower outlier zone. The Answer-Answer dataset is one of the most difficult datasets to predict for the state of the art methods. Nonetheless, even the BoW baseline reaches the IQR for this dataset. In the general state-of-the-art correlation (\texttt{ALL}), WISSE is located barely beyond the first quartile; in fact within the upper whisker of the box plot. 

\section{Discussion}\label{sec:9discussion}
The differences in the behavior of our sentence representation method with respect to the textual properties of each dataset are meaningful. As confirmed by our results, different genres of text give different results. 

The fact that WISSE outperformed the state of the art on the SICK dataset is interesting. This is because such a dataset constitutes a careful selection of sentence pairs coming from varied STS datasets (2012-2014). This selection included cross-level tasks \citep{jurgens2016cross}, which are the most difficult tasks in existent STS benchmarks. In fact, we observed that it is very difficult to surpass the $0.7200$ barrier and we think that it is mainly due to cross-level samples. These kind of samples are mostly used for textual entailment tasks, rather than for learning sentence representations. Nonetheless, supervised methods have been successful in learning generalization of both kind of tasks simultaneously \citep{yin2016abcnn}. There are now 5 sentence embedding methods within such a barrier (Glove+WR, Sent2vec, FastSent, C-PHRASE and WISSE). All these methods are completely based on neural networks. Our method (WISSE) integrates neural networks and Shannon's entropy, the former for learning word embeddings, and the second for weighting word embeddings. 

In addition, the experiments on diverse text properties presented by the SemEval (2016) dataset showed that WISSE offers the possibility to be prepared for varied scenarios. Such preparation consists of using proper hyperparameters for text properties, which is possible due to the modularity of our model. 

Shannon's entropy of words in the corpus can be used for weighting the contribution of each word embedding. Therefore, these two elements (words and their information contributions) are identifiable at the time that our method builds sentence representations. This is a partial\footnote{Partial identifiability is referred to as the fact that we are using word embeddings that do not come necessarily from an identifiable model.} identifiability allowing for future studies on statistical properties of sentence meaning. 

A surprising aspect of our approach is that the entropy is indeed a real number (a scalar). This fact differs considerably from purely neural network-based approaches: a whole matrix should be learned in order to capture interactions between word embeddings comprising a sentence representation. Thus our experiments were performed at relatively low computational cost. 

For each dataset we wanted to know whether our model is actually competitive with respect to the overall STS state of the art. We observed in Section \ref{sec:3distictionBrepresentationsAndSTS} that there is a wide variety of STS systems. Several of them are based on knowledge and language resources as well as on manually annotated similarity gold standards for supervised learning. As we reported in Section \ref{sec:semevalPosition}, these advantages do not always represent for a significant advantage over unsupervised sentence representation methods. In this sense, in most cases presented in this paper, WISSE succeed in dealing with the uncertainty of the mean correlations reached by 95\% of all state-of-the-art systems. This holds even when the variability of the overall performances is high. 

We think that the aforementioned facts indicate that WISSE is a competitive method for representing meaning in sentences.

There are datasets where most methods in the state of the art (including WISSE) reached low performances, i.e. FNWN and Answer-Answer. For the case of WISSE, we think this was because of the magnitude that the sentence representations attained due to the contributions of weighted word embeddings. It subsumes the word embeddings associated with the words composing a text snippet. The larger the number of words contained within, the larger the magnitude reached by the resulting representation\footnote{This affirmation is relative. In $\ell_p$ spaces, with $1<p<\infty$, it holds that $\|x\|_p=\left(\sum_{i=1}^\infty |x_i|^{p}\right)^{\frac{1}{p}} <\infty$.}. The FNWN and Answer-Answer datasets constitute cross-level STS tasks \citep{jurgens2016cross}, e.g. the comparison between a phrase and a paragraph. The length imbalance makes WISSE producing significantly different vectors to be compared. For instance, even when the meanings represented are very similar, length imbalance can cause meaningful geometric differences. This fact was confirmed as the cosine similarity was the best similarity function for the datasets in question (Section \ref{sec:resultsTuningSemeval}). In these cases we think that there were more useful similarity cues encoded in the angles of vectors than into their lengths (which are actually omitted by the cosine similarity). Conversely, for balanced length between compared snippets the metrics led to better results because such metrics exploited the fact that sentence meaning was encoded into the sentence representation magnitudes (and implicitly into the angles).

\section{Possible improvements}\label{sec:10possibleImprovements}

\paragraph{Blinded information sources} Our current experiments only considered a basic version of our model using three sources of information provided by Shannon's entropy (TF--IDF). Nonetheless, as we can see in eqs. (\ref{eq:idfa}) and (\ref{eq:idfb}), the usual IDF weights make na\"ive assumptions in terms of the probability measures of words and sentences. We think these assumptions might circumvent important things. The most important is the actual probability measures of the stochastic processes underlying the different levels and hierarchies of contexts. Clearly, uniform distributions are not the way sentences are drawn from the communication process. In this sense, we think that more accurate empirical estimations should be performed in order to uncover more accurate models of stochastic processes underlying language \citep{vapnik1998}. These estimations will provide us more granular information sources, better weighting estimates and probably better sentence representations (of course, it will depend on the requirements of the addressed NLP task). 

Consider the case of \textit{word order universals} \citep{Desmond1977}, which is an information source we have left blinded for the current version of our model. A speaker emitting a sentence like \ref{sent:1} generally emphasizes the agent (\textit{Paula}) performing the action (\textit{to strike}). Conversely, in sentences like \ref{sent:2}, the speaker emphasizes the
experiencer (\textit{Nacho}). This is not a general rule, but it surely provides an information spectrum related to informativeness of noun order. 

\begin{enumerate}[resume]
\item \textit{Paula struck Nacho.}\label{sent:1}
\item \textit{Nacho was beaten by Paula.}\label{sent:2}
\end{enumerate}
Notice that self similarity and hierarchy of these patterns can be found also in more complex constructions, e.g., \textit{Paula struck Nacho because his offense was inadmissible}. In this example, the noun phrase \textit{his offense} in the subordinate clause is also emphasized by the speaker (with respect to the adjective \textit{inadmissible}). Nonetheless, for the whole sentence this emphasis is hierarchically less important than that put onto the agent in the main clause (\textit{Paula}).

\paragraph{Advantages and disadvantages}
At the moment, WISSE does not capture properly specific information patterns, like negation adverbs. For instance, similar representations of the sentences \ref{sentEg:condem} and \ref{sentEg:notcondem} are built:
\begin{enumerate}
\item \textit{Computers will not condemn humanity.} \label{sentEg:condem}
\item \textit{Computers will condemn humanity.} \label{sentEg:notcondem}
\end{enumerate}
This holds because the same adverb form (\textit{not}) is used many times when the speaker needs to negate any given fact. Thus, from the point of view of the information embedding that WISSE uses at this moment, this word is almost not informative. We show in Table \ref{tab:not} different weights for the word \textit{not} within two randomly picked sentences from our corpus: 
\begin{table}[!ht]
\centering
\caption{TF--IDF weights for the adverb form \textit{not} within two sentences}
\vspace{3mm}
\label{tab:not}
\begin{tabular}{p{4cm}|p{7cm}}
Sentence                                            & TF--IDF weights\\
\hline
\textit{The man jumping is not wearing a shirt.}           &  ``jumping'' ($0.6537$), ``wearing'' ($0.4753$), ``shirt'' ($0.5679$), ``not'' ($\textbf{0.1894}$) \\
\textit{A girl is close to a boy whose face is not shown.} & ``girl'' ($0.4258$), ``close'' ($0.3415$), ``boy'' ($0.4339$), ``face'' ($0.3833$),  ``not'' ($\textbf{0.1959}$) \\
\hline
\end{tabular}
\end{table}

In fact, negation is an open issue in STS research. Depending on the way we consider the TF (either binary or logarithmic or frequency), the weights will change. Nonetheless their entropy keeps relatively constant. 

Notice that we are clearly identifying specific linguistic items participating in our model. This is an advantage of WISSE over most sentence representation methods. It is possible to manipulate the contribution of specific linguistic items within sentences. For instance, we could be interested in detecting similarity of sentences with respect to a specific entity (a noun phrase). In this case, it is sufficient to reinforce the weights associated with the word embeddings representing such an entity \citep{badarinza2017syntactic}. 
This identifiability is not easy to attain by other state-of-the-art purely neural sentence embedding methods.

\section{Conclusions}\label{sec:conclusions}
 
We evaluated our approach on well-known STS benchmarks. Our method outperformed the usual baselines of sentence representation (BoW and word embedding average). Furthermore, the evaluation on the SICK dataset showed that our method outperformed the state of the art methods by reaching a correlation of $\rho=0.724$. This is encouraging because this dataset is overall a bit difficult for unsupervised sentence representation methods. In fact, the barrier of $0.72$ is difficult to overcome.  

WISSE also outperformed state-of-the-art methods in a couple of datasets (Answer-Answer and Plagiarism) of the well-known SemEval (2016) STS benchmark. Particularly for the Answer-Answer dataset our method surpassed the state of the art methods by about $10\%$, although this is one of the most challenging datasets. For the remaining datasets, our method's performance was close to the state of the art methods (within $2.7\%$).

We also compared WISSE against a wide variety of methods that participated in the SemEval STS competition (2016). Although such methods were not restricted to unsupervised or external knowledge-free methods, our model surpassed the mean performance. Furthermore, in most cases WISSE surpassed 95\% of the competitors and it was close (within $4\%$) to the best STS system (which uses supervised learning, i.e. support vector regression and deep learning, as well as a variety of external resources such as WordNet, Wikipedia and named entity recognizers). 

Our experiments confirmed our hypothesis that it is possible to well represent sentences by using the link between the contexts learned by word embeddings and the entropy of embedded words. We exploited the mentioned link to learn without supervision the weights of a series of word embeddings representing a sentence. Interestingly, such weights are simple scalars that allowed our model to reach state-of-the-art performance in difficult STS tasks at low computational cost. 

The modularity of our model gives the possibility of configuring our model according to the properties of text, which allowed us to obtain the best performance in important STS tasks.
It is also interesting that such performances were obtained by using simple similarity functions (cosine, Euclidean, Manhattan). 

Finally, the low computational cost of our method makes it especially useful when only unlabeled text is available for learning sentence representations. Furthermore, our model is suitable for low-resource applications and for leveraging real-time and online analysis and applications of sentence representations.

\section*{Acknowledgements}
We thank to: CONACyT (Grant 386128), \textit{Laboratorio de c\'omputo de alto rendimiento} (IIMAS--UNAM), Alfredo Jos\'e Hern\'andez \'Alvarez (CCG--UNAM), The Computational Genomics Research Program (PGC/CCG--UNAM) and The Postgraduate Program in Computer Science and Engineering--UNAM. Also thanks to Sarah Elizabeth Campion (Qatar University) for helping others in an unselfish way.
\vspace*{-12pt}

\bibliography{elsarticle-template}

\begin{thebibliography}{78}
\expandafter\ifx\csname natexlab\endcsname\relax\def\natexlab#1{#1}\fi
\providecommand{\url}[1]{\texttt{#1}}
\providecommand{\href}[2]{#2}
\providecommand{\path}[1]{#1}
\providecommand{\DOIprefix}{doi:}
\providecommand{\ArXivprefix}{arXiv:}
\providecommand{\URLprefix}{URL: }
\providecommand{\Pubmedprefix}{pmid:}
\providecommand{\doi}[1]{\href{http://dx.doi.org/#1}{\path{#1}}}
\providecommand{\Pubmed}[1]{\href{pmid:#1}{\path{#1}}}
\providecommand{\bibinfo}[2]{#2}
\ifx\xfnm\undefined \def\xfnm[#1]{\unskip,\space#1}\fi
\bibitem[{Afzal et~al.(2016)Afzal, Wang and Liu}]{afzal2016mayonlp}
\bibinfo{author}{Afzal\xfnm[ N.]}, \bibinfo{author}{Wang\xfnm[ Y.]},
  \bibinfo{author}{Liu\xfnm[ H.]}.
\newblock \bibinfo{title}{Mayonlp at semeval-2016 task 1: Semantic textual
  similarity based on lexical semantic net and deep learning semantic model.}
\newblock In: \bibinfo{booktitle}{SemEval NAACL-HLT}. \bibinfo{year}{2016}. p.
  \bibinfo{pages}{674--679}.
\bibitem[{Agirre et~al.(2016)Agirre, Banea, Cer, Diab, Gonzalez-Agirre,
  Mihalcea, Rigau and Wiebe}]{agirre2016semeval}
\bibinfo{author}{Agirre\xfnm[ E.]}, \bibinfo{author}{Banea\xfnm[ C.]},
  \bibinfo{author}{Cer\xfnm[ D.]}, \bibinfo{author}{Diab\xfnm[ M.]},
  \bibinfo{author}{Gonzalez-Agirre\xfnm[ A.]}, \bibinfo{author}{Mihalcea\xfnm[
  R.]}, \bibinfo{author}{Rigau\xfnm[ G.]}, \bibinfo{author}{Wiebe\xfnm[ J.]}.
\newblock \bibinfo{title}{Semeval-2016 task 1: Semantic textual similarity,
  monolingual and cross-lingual evaluation}.
\newblock In: \bibinfo{booktitle}{Proceedings of SemEval}.
  \bibinfo{year}{2016}. p. \bibinfo{pages}{497--511}.
\bibitem[{Agirre et~al.(2013)Agirre, Cer, Diab, Gonzalez-Agirre and
  Guo}]{agirre2013sem}
\bibinfo{author}{Agirre\xfnm[ E.]}, \bibinfo{author}{Cer\xfnm[ D.]},
  \bibinfo{author}{Diab\xfnm[ M.]}, \bibinfo{author}{Gonzalez-Agirre\xfnm[
  A.]}, \bibinfo{author}{Guo\xfnm[ W.]}.
\newblock \bibinfo{title}{* sem 2013 shared task: Semantic textual similarity}.
\newblock In: \bibinfo{booktitle}{Second Joint Conference on Lexical and
  Computational Semantics (* SEM), Volume 1: Proceedings of the Main Conference
  and the Shared Task: Semantic Textual Similarity}.
  volume~\bibinfo{volume}{1}; \bibinfo{year}{2013}. p. \bibinfo{pages}{32--43}.
\bibitem[{Agirre et~al.(2012)Agirre, Diab, Cer and
  Gonzalez-Agirre}]{agirre2012semeval}
\bibinfo{author}{Agirre\xfnm[ E.]}, \bibinfo{author}{Diab\xfnm[ M.]},
  \bibinfo{author}{Cer\xfnm[ D.]}, \bibinfo{author}{Gonzalez-Agirre\xfnm[ A.]}.
\newblock \bibinfo{title}{Semeval-2012 task 6: A pilot on semantic textual
  similarity}.
\newblock In: \bibinfo{booktitle}{Proceedings of the First Joint Conference on
  Lexical and Computational Semantics-Volume 1: Proceedings of the main
  conference and the shared task, and Volume 2: Proceedings of the Sixth
  International Workshop on Semantic Evaluation}.
  \bibinfo{organization}{Association for Computational Linguistics};
  \bibinfo{year}{2012}. p. \bibinfo{pages}{385--393}.
\bibitem[{Aizawa(2003)}]{aizawa2003information}
\bibinfo{author}{Aizawa\xfnm[ A.]}.
\newblock \bibinfo{title}{An information-theoretic perspective of tf--idf
  measures}.
\newblock \bibinfo{journal}{Information Processing \& Management}
  \bibinfo{year}{2003};\bibinfo{volume}{39}(\bibinfo{number}{1}):\bibinfo{pages}{45--65}.
\bibitem[{Arora et~al.(2017)Arora, Liang and Ma}]{arora2017simple}
\bibinfo{author}{Arora\xfnm[ S.]}, \bibinfo{author}{Liang\xfnm[ Y.]},
  \bibinfo{author}{Ma\xfnm[ T.]}.
\newblock \bibinfo{title}{A simple but tough-to-beat baseline for sentence
  embeddings}.
\newblock \bibinfo{journal}{International Conference on Learning
  Representations (ICLR)} \bibinfo{year}{2017};.
\bibitem[{Arroyo-Fern{\'a}ndez(2015)}]{arroyo2015learning}
\bibinfo{author}{Arroyo-Fern{\'a}ndez\xfnm[ I.]}.
\newblock \bibinfo{title}{Learning kernels for semantic clustering: A deep
  approach}.
\newblock In: \bibinfo{booktitle}{NAACL-HLT 2015 Student Research Workshop
  (SRW)}. \bibinfo{year}{2015}. p. \bibinfo{pages}{79--87}.
\bibitem[{Arroyo-Fern\'{a}ndez and
  Meza~Ruiz(2017)}]{arroyofernandez-mezaruiz2017SemEval}
\bibinfo{author}{Arroyo-Fern\'{a}ndez\xfnm[ I.]},
  \bibinfo{author}{Meza~Ruiz\xfnm[ I.V.]}.
\newblock \bibinfo{title}{{LIPN-IIMAS} at {SemEval}-2017 task 1: Subword
  embeddings, attention recurrent neural networks and cross word alignment for
  semantic textual similarity}.
\newblock In: \bibinfo{booktitle}{Proceedings of the 11th International
  Workshop on Semantic Evaluation (SemEval-2017)}. \bibinfo{address}{Vancouver,
  Canada}: \bibinfo{publisher}{Association for Computational Linguistics
  (ACL)}; \bibinfo{year}{2017}. p. \bibinfo{pages}{199--203}.
\newblock \URLprefix \url{http://www.aclweb.org/anthology/S17-2031}.
\bibitem[{Arroyo-Fern\'andez et~al.(2016)Arroyo-Fern\'andez, Torres-Moreno,
  Sierra and Cabrera-Diego}]{arroyo-fernandez-kdir16}
\bibinfo{author}{Arroyo-Fern\'andez\xfnm[ I.]},
  \bibinfo{author}{Torres-Moreno\xfnm[ J.M.]}, \bibinfo{author}{Sierra\xfnm[
  G.]}, \bibinfo{author}{Cabrera-Diego\xfnm[ L.A.]}.
\newblock \bibinfo{title}{Automatic text summarization by non-topic relevance
  estimation}.
\newblock In: \bibinfo{booktitle}{Proceedings of the 8th International Joint
  Conference on Knowledge Discovery, Knowledge Engineering and Knowledge
  Management - Volume 1: KDIR,}. \bibinfo{year}{2016}. p.
  \bibinfo{pages}{89--100}.
\newblock \DOIprefix\doi{10.5220/0006053400890100}.
\bibitem[{Badarinza et~al.(2017)Badarinza, Sterca and
  Ionescu}]{badarinza2017syntactic}
\bibinfo{author}{Badarinza\xfnm[ I.]}, \bibinfo{author}{Sterca\xfnm[ A.I.]},
  \bibinfo{author}{Ionescu\xfnm[ M.]}.
\newblock \bibinfo{title}{Syntactic indexes for text retrieval}.
\newblock \bibinfo{journal}{Information Technology in Industry}
  \bibinfo{year}{2017};\bibinfo{volume}{5}:\bibinfo{pages}{24--28}.
\bibitem[{Baroni et~al.(2014)Baroni, Dinu and Kruszewski}]{baroni2014don}
\bibinfo{author}{Baroni\xfnm[ M.]}, \bibinfo{author}{Dinu\xfnm[ G.]},
  \bibinfo{author}{Kruszewski\xfnm[ G.]}.
\newblock \bibinfo{title}{Don't count, predict! a systematic comparison of
  context-counting vs. context-predicting semantic vectors.}
\newblock In: \bibinfo{booktitle}{ACL (1)}. \bibinfo{year}{2014}. p.
  \bibinfo{pages}{238--247}.
\bibitem[{Baroni and Lenci(2010)}]{baroni2010}
\bibinfo{author}{Baroni\xfnm[ M.]}, \bibinfo{author}{Lenci\xfnm[ A.]}.
\newblock \bibinfo{title}{Distributional memory: A general framework for
  corpus-based semantics}.
\newblock \bibinfo{journal}{Computational Linguistics}
  \bibinfo{year}{2010};\bibinfo{volume}{36}(\bibinfo{number}{4}):\bibinfo{pages}{673--721}.
\bibitem[{Bengio et~al.(2003)Bengio, Ducharme, Vincent and
  Jauvin}]{bengio2003neural}
\bibinfo{author}{Bengio\xfnm[ Y.]}, \bibinfo{author}{Ducharme\xfnm[ R.]},
  \bibinfo{author}{Vincent\xfnm[ P.]}, \bibinfo{author}{Jauvin\xfnm[ C.]}.
\newblock \bibinfo{title}{A neural probabilistic language model}.
\newblock \bibinfo{journal}{Journal of Machine Learning Research}
  \bibinfo{year}{2003};\bibinfo{volume}{3}(\bibinfo{number}{Feb}):\bibinfo{pages}{1137--1155}.
\bibitem[{Bentivogli et~al.(2016)Bentivogli, Bernardi, Marelli, Menini, Baroni
  and Zamparelli}]{bentivogli2016sick}
\bibinfo{author}{Bentivogli\xfnm[ L.]}, \bibinfo{author}{Bernardi\xfnm[ R.]},
  \bibinfo{author}{Marelli\xfnm[ M.]}, \bibinfo{author}{Menini\xfnm[ S.]},
  \bibinfo{author}{Baroni\xfnm[ M.]}, \bibinfo{author}{Zamparelli\xfnm[ R.]}.
\newblock \bibinfo{title}{Sick through the semeval glasses. lesson learned from
  the evaluation of compositional distributional semantic models on full
  sentences through semantic relatedness and textual entailment}.
\newblock \bibinfo{journal}{Language Resources and Evaluation}
  \bibinfo{year}{2016};\bibinfo{volume}{50}(\bibinfo{number}{1}):\bibinfo{pages}{95--124}.
\bibitem[{Bojanowski et~al.(2016)Bojanowski, Grave, Joulin and
  Mikolov}]{bojanowski2016enriching}
\bibinfo{author}{Bojanowski\xfnm[ P.]}, \bibinfo{author}{Grave\xfnm[ E.]},
  \bibinfo{author}{Joulin\xfnm[ A.]}, \bibinfo{author}{Mikolov\xfnm[ T.]}.
\newblock \bibinfo{title}{Enriching word vectors with subword information}.
\newblock \bibinfo{journal}{arXiv preprint arXiv:160704606}
  \bibinfo{year}{2016};.
\bibitem[{Brokos et~al.(2016)Brokos, Malakasiotis and
  Androutsopoulos}]{brokos2016using}
\bibinfo{author}{Brokos\xfnm[ G.I.]}, \bibinfo{author}{Malakasiotis\xfnm[ P.]},
  \bibinfo{author}{Androutsopoulos\xfnm[ I.]}.
\newblock \bibinfo{title}{Using centroids of word embeddings and word mover’s
  distance for biomedical document retrieval in question answering}.
\newblock \bibinfo{journal}{ACL 2016}
  \bibinfo{year}{2016};:\bibinfo{pages}{114}.
\bibitem[{Brychc{\i}n and Svoboda(2016)}]{brychcin2016uwb}
\bibinfo{author}{Brychc{\i}n\xfnm[ T.]}, \bibinfo{author}{Svoboda\xfnm[ L.]}.
\newblock \bibinfo{title}{Uwb at semeval-2016 task 1: Semantic textual
  similarity using lexical, syntactic, and semantic information}.
\newblock In: \bibinfo{booktitle}{Proceedings of SemEval}.
  \bibinfo{publisher}{ACL 2016}; \bibinfo{year}{2016}. p.
  \bibinfo{pages}{588--594}.
\bibitem[{Cer et~al.(2017)Cer, Diab, Agirre, Lopez-Gazpio and
  Specia}]{cerEtAl2017}
\bibinfo{author}{Cer\xfnm[ D.]}, \bibinfo{author}{Diab\xfnm[ M.]},
  \bibinfo{author}{Agirre\xfnm[ E.]}, \bibinfo{author}{Lopez-Gazpio\xfnm[ I.]},
  \bibinfo{author}{Specia\xfnm[ L.]}.
\newblock \bibinfo{title}{Semeval-2017 task 1: Semantic textual similarity
  multilingual and crosslingual focused evaluation}.
\newblock In: \bibinfo{booktitle}{Proceedings of the 11th International
  Workshop on Semantic Evaluation (SemEval-2017)}. \bibinfo{address}{Vancouver,
  Canada}: \bibinfo{publisher}{Association for Computational Linguistics};
  \bibinfo{year}{2017}. p. \bibinfo{pages}{1--14}.
\newblock \URLprefix \url{http://www.aclweb.org/anthology/S17-2001}.
\bibitem[{Charniak(1996)}]{charniak1996statistical}
\bibinfo{author}{Charniak\xfnm[ E.]}.
\newblock \bibinfo{title}{Statistical language learning}.
\newblock \bibinfo{publisher}{MIT Press}, \bibinfo{year}{1996}.
\bibitem[{Chen et~al.(2017)Chen, Xu, He and Wang}]{chen2017improving}
\bibinfo{author}{Chen\xfnm[ T.]}, \bibinfo{author}{Xu\xfnm[ R.]},
  \bibinfo{author}{He\xfnm[ Y.]}, \bibinfo{author}{Wang\xfnm[ X.]}.
\newblock \bibinfo{title}{Improving sentiment analysis via sentence type
  classification using bilstm-crf and cnn}.
\newblock \bibinfo{journal}{Expert Systems with Applications}
  \bibinfo{year}{2017};\bibinfo{volume}{72}:\bibinfo{pages}{221--230}.
\bibitem[{Collobert et~al.(2011)Collobert, Weston, Bottou, Karlen, Kavukcuoglu
  and Kuksa}]{collobert2011natural}
\bibinfo{author}{Collobert\xfnm[ R.]}, \bibinfo{author}{Weston\xfnm[ J.]},
  \bibinfo{author}{Bottou\xfnm[ L.]}, \bibinfo{author}{Karlen\xfnm[ M.]},
  \bibinfo{author}{Kavukcuoglu\xfnm[ K.]}, \bibinfo{author}{Kuksa\xfnm[ P.]}.
\newblock \bibinfo{title}{Natural language processing (almost) from scratch}.
\newblock \bibinfo{journal}{Journal of Machine Learning Research}
  \bibinfo{year}{2011};\bibinfo{volume}{12}(\bibinfo{number}{Aug}):\bibinfo{pages}{2493--2537}.
\bibitem[{De~Boom et~al.(2016)De~Boom, Van~Canneyt, Demeester and
  Dhoedt}]{de2016representation}
\bibinfo{author}{De~Boom\xfnm[ C.]}, \bibinfo{author}{Van~Canneyt\xfnm[ S.]},
  \bibinfo{author}{Demeester\xfnm[ T.]}, \bibinfo{author}{Dhoedt\xfnm[ B.]}.
\newblock \bibinfo{title}{Representation learning for very short texts using
  weighted word embedding aggregation}.
\newblock \bibinfo{journal}{Pattern Recognition Letters}
  \bibinfo{year}{2016};\bibinfo{volume}{80}:\bibinfo{pages}{150--156}.
\bibitem[{De~Marcken(1999)}]{de1999unsupervised}
\bibinfo{author}{De~Marcken\xfnm[ C.]}.
\newblock \bibinfo{title}{On the unsupervised induction of phrase-structure
  grammars}.
\newblock In: \bibinfo{booktitle}{Natural Language Processing Using Very Large
  Corpora}. \bibinfo{publisher}{Springer}; \bibinfo{year}{1999}. p.
  \bibinfo{pages}{191--208}.
\bibitem[{Derbyshire(1977)}]{Desmond1977}
\bibinfo{author}{Derbyshire\xfnm[ D.C.]}.
\newblock \bibinfo{title}{Word order universals and the existence of ovs
  languages}.
\newblock \bibinfo{journal}{Linguistic Inquiry}
  \bibinfo{year}{1977};\bibinfo{volume}{8}(\bibinfo{number}{3}):\bibinfo{pages}{590--599}.
\newblock \URLprefix \url{http://www.jstor.org/stable/4178003}.
\bibitem[{Elman(1991)}]{elman1991distributed}
\bibinfo{author}{Elman\xfnm[ J.L.]}.
\newblock \bibinfo{title}{Distributed representations, simple recurrent
  networks, and grammatical structure}.
\newblock \bibinfo{journal}{Machine Learning}
  \bibinfo{year}{1991};\bibinfo{volume}{7}(\bibinfo{number}{2-3}):\bibinfo{pages}{195--225}.
\bibitem[{Er et~al.(2016)Er, Zhang, Wang and Pratama}]{er2016attention}
\bibinfo{author}{Er\xfnm[ M.J.]}, \bibinfo{author}{Zhang\xfnm[ Y.]},
  \bibinfo{author}{Wang\xfnm[ N.]}, \bibinfo{author}{Pratama\xfnm[ M.]}.
\newblock \bibinfo{title}{Attention pooling-based convolutional neural network
  for sentence modelling}.
\newblock \bibinfo{journal}{Information Sciences}
  \bibinfo{year}{2016};\bibinfo{volume}{373}:\bibinfo{pages}{388--403}.
\bibitem[{Ferrero et~al.(2017)Ferrero, Besacier, Schwab and
  Agn\`{e}s}]{ferrero-EtAl:2017:SemEval}
\bibinfo{author}{Ferrero\xfnm[ J.]}, \bibinfo{author}{Besacier\xfnm[ L.]},
  \bibinfo{author}{Schwab\xfnm[ D.]}, \bibinfo{author}{Agn\`{e}s\xfnm[ F.]}.
\newblock \bibinfo{title}{Compilig at semeval-2017 task 1: Cross-language
  plagiarism detection methods for semantic textual similarity}.
\newblock In: \bibinfo{booktitle}{Proceedings of the 11th International
  Workshop on Semantic Evaluation (SemEval-2017)}. \bibinfo{address}{Vancouver,
  Canada}: \bibinfo{publisher}{Association for Computational Linguistics};
  \bibinfo{year}{2017}. p. \bibinfo{pages}{100--105}.
\newblock \URLprefix \url{http://www.aclweb.org/anthology/S17-2012}.
\bibitem[{Firth(1957)}]{Firth1957}
\bibinfo{author}{Firth\xfnm[ J.R.]}.
\newblock \bibinfo{title}{Papers in linguistics 1934-1951}.
\newblock \bibinfo{address}{London}: \bibinfo{publisher}{Oxford University
  Press}, \bibinfo{year}{1957}.
\bibitem[{Fourier(1822)}]{fourier1822theorie}
\bibinfo{author}{Fourier\xfnm[ J.]}.
\newblock \bibinfo{title}{Theorie analytique de la chaleur}.
\newblock \bibinfo{publisher}{Chez Firmin Didot, p{\`e}re et fils},
  \bibinfo{year}{1822}.
\bibitem[{Han et~al.(2013)Han, Kashyap, Finin, Mayfield and
  Weese}]{han2013umbc}
\bibinfo{author}{Han\xfnm[ L.]}, \bibinfo{author}{Kashyap\xfnm[ A.]},
  \bibinfo{author}{Finin\xfnm[ T.]}, \bibinfo{author}{Mayfield\xfnm[ J.]},
  \bibinfo{author}{Weese\xfnm[ J.]}.
\newblock \bibinfo{title}{Umbc ebiquity-core: Semantic textual similarity
  systems}.
\newblock \bibinfo{journal}{Atlanta, GA, USA}
  \bibinfo{year}{2013};\bibinfo{volume}{44}.
\bibitem[{Harris(1968)}]{harris1968}
\bibinfo{author}{Harris\xfnm[ Z.S.]}.
\newblock \bibinfo{title}{Mathematical Structures of Language}.
\newblock \bibinfo{address}{New York, NY, USA}: \bibinfo{publisher}{Wiley},
  \bibinfo{year}{1968}.
\bibitem[{Hatzivassiloglou et~al.(1999)Hatzivassiloglou, Klavans and
  Eskin}]{hatzivassiloglou1999detecting}
\bibinfo{author}{Hatzivassiloglou\xfnm[ V.]}, \bibinfo{author}{Klavans\xfnm[
  J.L.]}, \bibinfo{author}{Eskin\xfnm[ E.]}.
\newblock \bibinfo{title}{Detecting text similarity over short passages:
  Exploring linguistic feature combinations via machine learning}.
\newblock In: \bibinfo{booktitle}{Proceedings of the 1999 joint sigdat
  conference on empirical methods in natural language processing and very large
  corpora}. \bibinfo{year}{1999}. p. \bibinfo{pages}{203--212}.
\bibitem[{Hill et~al.(2016)Hill, Cho and Korhonen}]{hill2016learning}
\bibinfo{author}{Hill\xfnm[ F.]}, \bibinfo{author}{Cho\xfnm[ K.]},
  \bibinfo{author}{Korhonen\xfnm[ A.]}.
\newblock \bibinfo{title}{Learning distributed representations of sentences
  from unlabelled data}.
\newblock In: \bibinfo{booktitle}{Proceedings of NAACL-HLT}.
  \bibinfo{year}{2016}. p. \bibinfo{pages}{1367--1377}.
\bibitem[{Hinton et~al.(1986)Hinton, McClelland and
  Rumelhart}]{hinton1986distributed}
\bibinfo{author}{Hinton\xfnm[ G.]}, \bibinfo{author}{McClelland\xfnm[ J.]},
  \bibinfo{author}{Rumelhart\xfnm[ D.]}.
\newblock \bibinfo{title}{Distributed representations}.
\newblock In: \bibinfo{booktitle}{Parallel distributed processing: explorations
  in the microstructure of cognition, vol. 1}. \bibinfo{organization}{MIT
  Press}; \bibinfo{year}{1986}. p. \bibinfo{pages}{77--109}.
\bibitem[{Ji and Eisenstein(2013)}]{ji2013discriminative}
\bibinfo{author}{Ji\xfnm[ Y.]}, \bibinfo{author}{Eisenstein\xfnm[ J.]}.
\newblock \bibinfo{title}{Discriminative improvements to distributional
  sentence similarity.}
\newblock In: \bibinfo{booktitle}{EMNLP}. \bibinfo{year}{2013}. p.
  \bibinfo{pages}{891--896}.
\bibitem[{Jurgens et~al.(2016)Jurgens, Pilehvar and Navigli}]{jurgens2016cross}
\bibinfo{author}{Jurgens\xfnm[ D.]}, \bibinfo{author}{Pilehvar\xfnm[ M.T.]},
  \bibinfo{author}{Navigli\xfnm[ R.]}.
\newblock \bibinfo{title}{Cross level semantic similarity: an evaluation
  framework for universal measures of similarity}.
\newblock \bibinfo{journal}{Language Resources and Evaluation}
  \bibinfo{year}{2016};\bibinfo{volume}{50}(\bibinfo{number}{1}):\bibinfo{pages}{5--33}.
\bibitem[{Kalchbrenner et~al.(2014)Kalchbrenner, Grefenstette and
  Blunsom}]{Kalchbrenner2014}
\bibinfo{author}{Kalchbrenner\xfnm[ N.]}, \bibinfo{author}{Grefenstette\xfnm[
  E.]}, \bibinfo{author}{Blunsom\xfnm[ P.]}.
\newblock \bibinfo{title}{A convolutional neural network for modelling
  sentences}.
\newblock \bibinfo{journal}{Proceedings of the 52nd Annual Meeting of the
  Association for Computational Linguistics} \bibinfo{year}{2014};\URLprefix
  \url{http://goo.gl/EsQCuC}.
\bibitem[{Kenter and de~Rijke(2015)}]{kenter2015short}
\bibinfo{author}{Kenter\xfnm[ T.]}, \bibinfo{author}{de~Rijke\xfnm[ M.]}.
\newblock \bibinfo{title}{Short text similarity with word embeddings}.
\newblock In: \bibinfo{booktitle}{Proceedings of the 24th ACM International on
  Conference on Information and Knowledge Management}.
  \bibinfo{organization}{ACM}; \bibinfo{year}{2015}. p.
  \bibinfo{pages}{1411--1420}.
\bibitem[{King et~al.(2016)King, Gharbieh, Park and Cook}]{king2016unbnlp}
\bibinfo{author}{King\xfnm[ M.]}, \bibinfo{author}{Gharbieh\xfnm[ W.]},
  \bibinfo{author}{Park\xfnm[ S.]}, \bibinfo{author}{Cook\xfnm[ P.]}.
\newblock \bibinfo{title}{Unbnlp at semeval-2016 task 1: Semantic textual
  similarity: A unified framework for semantic processing and evaluation}.
\newblock \bibinfo{journal}{Proceedings of SemEval}
  \bibinfo{year}{2016};:\bibinfo{pages}{732--735}.
\bibitem[{Kintsch and Mangalath(2011)}]{kintsch2011construction}
\bibinfo{author}{Kintsch\xfnm[ W.]}, \bibinfo{author}{Mangalath\xfnm[ P.]}.
\newblock \bibinfo{title}{The construction of meaning}.
\newblock \bibinfo{journal}{Topics in Cognitive Science}
  \bibinfo{year}{2011};\bibinfo{volume}{3}(\bibinfo{number}{2}):\bibinfo{pages}{346--370}.
\bibitem[{Kiros et~al.(2015)Kiros, Zhu, Salakhutdinov, Zemel, Urtasun, Torralba
  and Fidler}]{kiros2015skip}
\bibinfo{author}{Kiros\xfnm[ R.]}, \bibinfo{author}{Zhu\xfnm[ Y.]},
  \bibinfo{author}{Salakhutdinov\xfnm[ R.R.]}, \bibinfo{author}{Zemel\xfnm[
  R.]}, \bibinfo{author}{Urtasun\xfnm[ R.]}, \bibinfo{author}{Torralba\xfnm[
  A.]}, \bibinfo{author}{Fidler\xfnm[ S.]}.
\newblock \bibinfo{title}{Skip-thought vectors}.
\newblock In: \bibinfo{booktitle}{Advances in neural information processing
  systems}. \bibinfo{year}{2015}. p. \bibinfo{pages}{3294--3302}.
\bibitem[{Landauer et~al.(1998)Landauer, Foltz and Laham}]{landauer1998}
\bibinfo{author}{Landauer\xfnm[ T.K.]}, \bibinfo{author}{Foltz\xfnm[ P.W.]},
  \bibinfo{author}{Laham\xfnm[ D.]}.
\newblock \bibinfo{title}{An introduction to latent semantic analysis}.
\newblock \bibinfo{journal}{Discourse Processes}
  \bibinfo{year}{1998};\bibinfo{volume}{25}(\bibinfo{number}{2-3}):\bibinfo{pages}{259--284}.
\bibitem[{Le and Mikolov(2014)}]{mikolov2014prhase}
\bibinfo{author}{Le\xfnm[ Q.]}, \bibinfo{author}{Mikolov\xfnm[ T.]}.
\newblock \bibinfo{title}{Distributed representations of sentences and
  documents}.
\newblock In: \bibinfo{booktitle}{31st International Conference on Machine
  Learning, {ICML} 2014, Beijing, China, 21-26 June}. \bibinfo{year}{2014}. p.
  \bibinfo{pages}{1188--1196}.
\newblock \URLprefix \url{http://jmlr.org/proceedings/papers/v32/le14.html}.
\bibitem[{Levy and Goldberg(2014)}]{levy2014dependency}
\bibinfo{author}{Levy\xfnm[ O.]}, \bibinfo{author}{Goldberg\xfnm[ Y.]}.
\newblock \bibinfo{title}{Dependency-based word embeddings.}
\newblock In: \bibinfo{booktitle}{ACL (2)}. \bibinfo{year}{2014}. p.
  \bibinfo{pages}{302--308}.
\bibitem[{Manning et~al.(2009)Manning, Raghavan and Sch\"utze}]{manning2009}
\bibinfo{author}{Manning\xfnm[ C.D.]}, \bibinfo{author}{Raghavan\xfnm[ P.]},
  \bibinfo{author}{Sch\"utze\xfnm[ H.]}.
\newblock \bibinfo{title}{An Introduction to Information Retrieval}.
\newblock \bibinfo{address}{Cambridge, United Kingdom}:
  \bibinfo{publisher}{Cambridge University Press}, \bibinfo{year}{2009}.
\bibitem[{Martin and Berry(2007)}]{martin2007mathematical}
\bibinfo{author}{Martin\xfnm[ D.I.]}, \bibinfo{author}{Berry\xfnm[ M.W.]}.
\newblock \bibinfo{title}{Mathematical foundations behind latent semantic
  analysis}.
\newblock \bibinfo{journal}{Handbook of latent semantic analysis}
  \bibinfo{year}{2007};:\bibinfo{pages}{35--56}.
\bibitem[{Meza-Ruiz and Riedel(2009)}]{meza2009jointly}
\bibinfo{author}{Meza-Ruiz\xfnm[ I.]}, \bibinfo{author}{Riedel\xfnm[ S.]}.
\newblock \bibinfo{title}{Jointly identifying predicates, arguments and senses
  using markov logic}.
\newblock In: \bibinfo{booktitle}{Proceedings of Human Language Technologies:
  The 2009 Annual Conference of the North American Chapter of the Association
  for Computational Linguistics}. \bibinfo{organization}{Association for
  Computational Linguistics}; \bibinfo{year}{2009}. p.
  \bibinfo{pages}{155--163}.
\bibitem[{Mihalcea et~al.(2006)Mihalcea, Corley, Strapparava
  et~al.}]{mihalcea2006corpus}
\bibinfo{author}{Mihalcea\xfnm[ R.]}, \bibinfo{author}{Corley\xfnm[ C.]},
  \bibinfo{author}{Strapparava\xfnm[ C.]}, et~al.
\newblock \bibinfo{title}{Corpus-based and knowledge-based measures of text
  semantic similarity}.
\newblock In: \bibinfo{booktitle}{AAAI}. volume~\bibinfo{volume}{6};
  \bibinfo{year}{2006}. p. \bibinfo{pages}{775--780}.
\bibitem[{Mikolov et~al.(2013{\natexlab{a}})Mikolov, Chen, Corrado and
  Dean}]{mikolov2013efficient}
\bibinfo{author}{Mikolov\xfnm[ T.]}, \bibinfo{author}{Chen\xfnm[ K.]},
  \bibinfo{author}{Corrado\xfnm[ G.]}, \bibinfo{author}{Dean\xfnm[ J.]}.
\newblock \bibinfo{title}{Efficient estimation of word representations in
  vector space}.
\newblock \bibinfo{journal}{arXiv preprint arXiv:13013781}
  \bibinfo{year}{2013}{\natexlab{a}};.
\bibitem[{Mikolov et~al.(2013{\natexlab{b}})Mikolov, Sutskever, Chen, Corrado
  and Dean}]{mikolov2013distributed}
\bibinfo{author}{Mikolov\xfnm[ T.]}, \bibinfo{author}{Sutskever\xfnm[ I.]},
  \bibinfo{author}{Chen\xfnm[ K.]}, \bibinfo{author}{Corrado\xfnm[ G.S.]},
  \bibinfo{author}{Dean\xfnm[ J.]}.
\newblock \bibinfo{title}{Distributed representations of words and phrases and
  their compositionality}.
\newblock In: \bibinfo{booktitle}{Advances in Neural Information Processing
  Systems}. \bibinfo{year}{2013}{\natexlab{b}}. p. \bibinfo{pages}{3111--3119}.
\bibitem[{Mitchell and Lapata(2010)}]{lapata2010}
\bibinfo{author}{Mitchell\xfnm[ J.]}, \bibinfo{author}{Lapata\xfnm[ M.]}.
\newblock \bibinfo{title}{Composition in distributional models of semantics}.
\newblock \bibinfo{journal}{Cognitive Science}
  \bibinfo{year}{2010};\bibinfo{volume}{34}(\bibinfo{number}{34}):\bibinfo{pages}{1388--1429}.
\newblock \bibinfo{note}{Cognitive Science Society, ISSN: 1551-6709}.
\bibitem[{Mitra et~al.(1997)Mitra, Buckley, Singhal and
  Cardie}]{mitra1997analysis}
\bibinfo{author}{Mitra\xfnm[ M.]}, \bibinfo{author}{Buckley\xfnm[ C.]},
  \bibinfo{author}{Singhal\xfnm[ A.]}, \bibinfo{author}{Cardie\xfnm[ C.]}.
\newblock \bibinfo{title}{An analysis of statistical and syntactic phrases}.
\newblock In: \bibinfo{booktitle}{Computer-Assisted Information Searching on
  Internet}. \bibinfo{organization}{Le Centre de Hautes Etudes Internationales
  d'Informatique Documentaire}; \bibinfo{year}{1997}. p.
  \bibinfo{pages}{200--214}.
\bibitem[{Onan et~al.(2017)Onan, Koruko{\u{g}}lu and Bulut}]{onan2017hybrid}
\bibinfo{author}{Onan\xfnm[ A.]}, \bibinfo{author}{Koruko{\u{g}}lu\xfnm[ S.]},
  \bibinfo{author}{Bulut\xfnm[ H.]}.
\newblock \bibinfo{title}{A hybrid ensemble pruning approach based on consensus
  clustering and multi-objective evolutionary algorithm for sentiment
  classification}.
\newblock \bibinfo{journal}{Information Processing \& Management}
  \bibinfo{year}{2017};\bibinfo{volume}{53}(\bibinfo{number}{4}):\bibinfo{pages}{814--833}.
\bibitem[{Osteyee and Good(1974)}]{Osteyee1974}
\bibinfo{author}{Osteyee\xfnm[ D.B.]}, \bibinfo{author}{Good\xfnm[ I.J.]}.
\newblock \bibinfo{title}{Expected mutual information}.
\newblock In: \bibinfo{booktitle}{Information, Weight of Evidence, the
  Singularity between Probability Measures and Signal Detection}.
  \bibinfo{address}{Berlin, Germany}: \bibinfo{publisher}{Springer};
  \bibinfo{year}{1974}. p. \bibinfo{pages}{26--38}.
\newblock \DOIprefix\doi{10.1007/BFb0064132}.
\bibitem[{Pagliardini et~al.(2017)Pagliardini, Gupta and Jaggi}]{pgj2017unsup}
\bibinfo{author}{Pagliardini\xfnm[ M.]}, \bibinfo{author}{Gupta\xfnm[ P.]},
  \bibinfo{author}{Jaggi\xfnm[ M.]}.
\newblock \bibinfo{title}{{Unsupervised Learning of Sentence Embeddings using
  Compositional n-Gram Features}}.
\newblock \bibinfo{journal}{arXiv}
  \bibinfo{year}{2017};\href{http://arxiv.org/abs/1703.02507}{\tt
  arXiv:1703.02507}.
\bibitem[{Pennington et~al.(2014)Pennington, Socher and
  Manning}]{pennington2014glove}
\bibinfo{author}{Pennington\xfnm[ J.]}, \bibinfo{author}{Socher\xfnm[ R.]},
  \bibinfo{author}{Manning\xfnm[ C.D.]}.
\newblock \bibinfo{title}{Glove: Global vectors for word representation}.
\newblock In: \bibinfo{booktitle}{Empirical Methods in Natural Language
  Processing (EMNLP)}. \bibinfo{year}{2014}. p. \bibinfo{pages}{1532--1543}.
\newblock \URLprefix \url{http://www.aclweb.org/anthology/D14-1162}.
\bibitem[{Pereira(2000)}]{pereira2000formal}
\bibinfo{author}{Pereira\xfnm[ F.]}.
\newblock \bibinfo{title}{Formal grammar and information theory: together
  again?}
\newblock \bibinfo{journal}{Philosophical Transactions of the Royal Society of
  London A: Mathematical, Physical and Engineering Sciences}
  \bibinfo{year}{2000};\bibinfo{volume}{358}(\bibinfo{number}{1769}):\bibinfo{pages}{1239--1253}.
\bibitem[{Pham et~al.(2015)Pham, Kruszewski, Lazaridou and
  Baroni}]{nghia2015jointly}
\bibinfo{author}{Pham\xfnm[ N.T.]}, \bibinfo{author}{Kruszewski\xfnm[ G.]},
  \bibinfo{author}{Lazaridou\xfnm[ A.]}, \bibinfo{author}{Baroni\xfnm[ M.]}.
\newblock \bibinfo{title}{Jointly optimizing word representations for lexical
  and sentential tasks with the c-phrase model.}
\newblock In: \bibinfo{booktitle}{ACL (1)}. \bibinfo{year}{2015}. p.
  \bibinfo{pages}{971--981}.
\bibitem[{Pilehvar and Navigli(2015)}]{PILEHVAR201595}
\bibinfo{author}{Pilehvar\xfnm[ M.T.]}, \bibinfo{author}{Navigli\xfnm[ R.]}.
\newblock \bibinfo{title}{From senses to texts: An all-in-one graph-based
  approach for measuring semantic similarity}.
\newblock \bibinfo{journal}{Artificial Intelligence}
  \bibinfo{year}{2015};\bibinfo{volume}{228}:\bibinfo{pages}{95 -- 128}.
\bibitem[{Robertson(2004)}]{robertson2004}
\bibinfo{author}{Robertson\xfnm[ S.]}.
\newblock \bibinfo{title}{Understanding inverse document frequency: on
  theoretical arguments for idf}.
\newblock \bibinfo{journal}{Journal of Documentation}
  \bibinfo{year}{2004};\bibinfo{volume}{60}(\bibinfo{number}{5}):\bibinfo{pages}{503--520}.
\bibitem[{Rong(2014)}]{Rong14}
\bibinfo{author}{Rong\xfnm[ X.]}.
\newblock \bibinfo{title}{word2vec parameter learning explained}.
\newblock \bibinfo{journal}{CoRR}
  \bibinfo{year}{2014};\bibinfo{volume}{abs/1411.2738}.
\newblock \URLprefix \url{http://arxiv.org/abs/1411.2738}.
\bibitem[{Rychalska et~al.(2016)Rychalska, Pakulska, Chodorowska, Walczak and
  Andruszkiewicz}]{rychalska2016samsung}
\bibinfo{author}{Rychalska\xfnm[ B.]}, \bibinfo{author}{Pakulska\xfnm[ K.]},
  \bibinfo{author}{Chodorowska\xfnm[ K.]}, \bibinfo{author}{Walczak\xfnm[ W.]},
  \bibinfo{author}{Andruszkiewicz\xfnm[ P.]}.
\newblock \bibinfo{title}{Samsung {Poland NLP Team at SemEval-2016 Task 1}:
  {Necessity} for diversity; combining recursive autoencoders, {Wordnet} and
  ensemble methods to measure semantic similarity}.
\newblock In: \bibinfo{booktitle}{Proceedings of the 10th International
  Workshop on Semantic Evaluation (SemEval 2016), San Diego, CA, USA}.
  \bibinfo{year}{2016}. p. \bibinfo{pages}{602--608}.
\bibitem[{Salton and Buckley(1988)}]{salton1988term}
\bibinfo{author}{Salton\xfnm[ G.]}, \bibinfo{author}{Buckley\xfnm[ C.]}.
\newblock \bibinfo{title}{Term-weighting approaches in automatic text
  retrieval}.
\newblock \bibinfo{journal}{Information Processing \& Management}
  \bibinfo{year}{1988};\bibinfo{volume}{24}(\bibinfo{number}{5}):\bibinfo{pages}{513--523}.
\bibitem[{Salton et~al.(1983)Salton, Fox and Wu}]{salton1983extended}
\bibinfo{author}{Salton\xfnm[ G.]}, \bibinfo{author}{Fox\xfnm[ E.A.]},
  \bibinfo{author}{Wu\xfnm[ H.]}.
\newblock \bibinfo{title}{Extended {Boolean} information retrieval}.
\newblock \bibinfo{journal}{Communications of the ACM}
  \bibinfo{year}{1983};\bibinfo{volume}{26}(\bibinfo{number}{11}):\bibinfo{pages}{1022--1036}.
\bibitem[{Sch{\"o}lkopf et~al.(1997)Sch{\"o}lkopf, Simard, Smola and
  Vapnik}]{scholkopf1997prior}
\bibinfo{author}{Sch{\"o}lkopf\xfnm[ B.]}, \bibinfo{author}{Simard\xfnm[ P.]},
  \bibinfo{author}{Smola\xfnm[ A.]}, \bibinfo{author}{Vapnik\xfnm[ V.]}.
\newblock \bibinfo{title}{Prior knowledge in support vector kernels}.
\newblock In: \bibinfo{booktitle}{Proceedings of the 10th International
  Conference on Neural Information Processing Systems}.
  \bibinfo{address}{Cambridge, MA}: \bibinfo{organization}{MIT Press};
  \bibinfo{year}{1997}. p. \bibinfo{pages}{640--646}.
\bibitem[{Shannon(1949)}]{shannon1949communication}
\bibinfo{author}{Shannon\xfnm[ C.E.]}.
\newblock \bibinfo{title}{Communication theory of secrecy systems*}.
\newblock \bibinfo{journal}{Bell system technical journal}
  \bibinfo{year}{1949};\bibinfo{volume}{28}(\bibinfo{number}{4}):\bibinfo{pages}{656--715}.
\bibitem[{Sp\"ark~Jones(1972)}]{sparkJones1972}
\bibinfo{author}{Sp\"ark~Jones\xfnm[ K.]}.
\newblock \bibinfo{title}{A statistical interpretation of term specificity and
  its application in retrieval}.
\newblock \bibinfo{journal}{Journal of Documentation}
  \bibinfo{year}{1972};\bibinfo{volume}{28}(\bibinfo{number}{1}):\bibinfo{pages}{11--21}.
\bibitem[{Sultan et~al.(2014)Sultan, Bethard and Sumner}]{sultan2014back}
\bibinfo{author}{Sultan\xfnm[ M.A.]}, \bibinfo{author}{Bethard\xfnm[ S.]},
  \bibinfo{author}{Sumner\xfnm[ T.]}.
\newblock \bibinfo{title}{Back to basics for monolingual alignment: Exploiting
  word similarity and contextual evidence}.
\newblock \bibinfo{journal}{Transactions of the Association for Computational
  Linguistics}
  \bibinfo{year}{2014};\bibinfo{volume}{2}:\bibinfo{pages}{219--230}.
\bibitem[{T{\v{e}}{\v{s}}it{\v{e}}lov{\'a}(1992)}]{tesitelova1992}
\bibinfo{author}{T{\v{e}}{\v{s}}it{\v{e}}lov{\'a}\xfnm[ M.]}.
\newblock \bibinfo{title}{Quantitative linguistics}.
\newblock Linguistics and literary studies in Eastern Europe.
  \bibinfo{publisher}{Academia Publishing House of the Czechoslovak Academy of
  Sciences}, \bibinfo{year}{1992}.
\bibitem[{Tian et~al.(2017)Tian, Okazaki and Inui}]{Tian2017}
\bibinfo{author}{Tian\xfnm[ R.]}, \bibinfo{author}{Okazaki\xfnm[ N.]},
  \bibinfo{author}{Inui\xfnm[ K.]}.
\newblock \bibinfo{title}{The mechanism of additive composition}.
\newblock \bibinfo{journal}{Machine Learning}
  \bibinfo{year}{2017};\bibinfo{volume}{106}(\bibinfo{number}{7}):\bibinfo{pages}{1083--1130}.
\bibitem[{Vapnik(1998)}]{vapnik1998}
\bibinfo{author}{Vapnik\xfnm[ V.N.]}.
\newblock \bibinfo{title}{Statistical Learning Theory}.
\newblock \bibinfo{address}{New York, NY}: \bibinfo{publisher}{Wiley},
  \bibinfo{year}{1998}.
\bibitem[{Wieting et~al.(2016)Wieting, Bansal, Gimpel and
  Livescu}]{wieting-EtAl:2016:EMNLP2016}
\bibinfo{author}{Wieting\xfnm[ J.]}, \bibinfo{author}{Bansal\xfnm[ M.]},
  \bibinfo{author}{Gimpel\xfnm[ K.]}, \bibinfo{author}{Livescu\xfnm[ K.]}.
\newblock \bibinfo{title}{Charagram: Embedding words and sentences via
  character n-grams}.
\newblock In: \bibinfo{booktitle}{Proceedings of the 2016 Conference on
  Empirical Methods in Natural Language Processing}. \bibinfo{address}{Austin,
  Texas}: \bibinfo{publisher}{Association for Computational Linguistics};
  \bibinfo{year}{2016}. p. \bibinfo{pages}{1504--1515}.
\newblock \URLprefix \url{https://aclweb.org/anthology/D16-1157}.
\bibitem[{Yazdani and Popescu-Belis(2013)}]{yazdani2013computing}
\bibinfo{author}{Yazdani\xfnm[ M.]}, \bibinfo{author}{Popescu-Belis\xfnm[ A.]}.
\newblock \bibinfo{title}{Computing text semantic relatedness using the
  contents and links of a hypertext encyclopedia}.
\newblock \bibinfo{journal}{Artificial Intelligence}
  \bibinfo{year}{2013};\bibinfo{volume}{194}:\bibinfo{pages}{176--202}.
\bibitem[{Yin and Sch{\"{u}tze}(2015)}]{Yin2015DiscriminativePE}
\bibinfo{author}{Yin\xfnm[ W.]}, \bibinfo{author}{Sch{\"{u}tze}\xfnm[ H.]}.
\newblock \bibinfo{title}{Discriminative phrase embedding for paraphrase
  identification}.
\newblock In: \bibinfo{booktitle}{Proceedings of HLT-NAACL}.
  \bibinfo{year}{2015}. p. \bibinfo{pages}{1368--1373}.
\bibitem[{Yin et~al.(2016)Yin, Sch{\"u}tze, Xiang and Zhou}]{yin2016abcnn}
\bibinfo{author}{Yin\xfnm[ W.]}, \bibinfo{author}{Sch{\"u}tze\xfnm[ H.]},
  \bibinfo{author}{Xiang\xfnm[ B.]}, \bibinfo{author}{Zhou\xfnm[ B.]}.
\newblock \bibinfo{title}{Abcnn: Attention-based convolutional neural network
  for modeling sentence pairs}.
\newblock \bibinfo{journal}{Transactions of the Association for Computational
  Linguistics}
  \bibinfo{year}{2016};\bibinfo{volume}{4}:\bibinfo{pages}{259--272}.
\bibitem[{Yu et~al.(2017)Yu, Xie, Xiao and Chng}]{yu2017learning}
\bibinfo{author}{Yu\xfnm[ J.]}, \bibinfo{author}{Xie\xfnm[ L.]},
  \bibinfo{author}{Xiao\xfnm[ X.]}, \bibinfo{author}{Chng\xfnm[ E.S.]}.
\newblock \bibinfo{title}{Learning distributed sentence representations for
  story segmentation}.
\newblock \bibinfo{journal}{Signal Processing} \bibinfo{year}{2017};.
\bibitem[{Zhang et~al.(2012)Zhang, Ge and He}]{zhang2012mutual}
\bibinfo{author}{Zhang\xfnm[ Z.]}, \bibinfo{author}{Ge\xfnm[ S.S.]},
  \bibinfo{author}{He\xfnm[ H.]}.
\newblock \bibinfo{title}{Mutual-reinforcement document summarization using
  embedded graph based sentence clustering for storytelling}.
\newblock \bibinfo{journal}{Information Processing \& Management}
  \bibinfo{year}{2012};\bibinfo{volume}{48}(\bibinfo{number}{4}):\bibinfo{pages}{767--778}.
\bibitem[{Zheng and Callan(2015)}]{zheng2015learning}
\bibinfo{author}{Zheng\xfnm[ G.]}, \bibinfo{author}{Callan\xfnm[ J.]}.
\newblock \bibinfo{title}{Learning to reweight terms with distributed
  representations}.
\newblock In: \bibinfo{booktitle}{Proceedings of the 38th International ACM
  SIGIR Conference on Research and Development in Information Retrieval}.
  \bibinfo{organization}{ACM}; \bibinfo{year}{2015}. p.
  \bibinfo{pages}{575--584}.

\end{thebibliography}

\end{document}